\documentclass[runningheads]{llncs}

\usepackage[mobile]{eccv}

\usepackage{eccvabbrv}
\usepackage{colortbl}
\usepackage{rotate}
\usepackage{graphicx}
\usepackage{booktabs}
\usepackage{multirow}
\usepackage{amsmath, bm}
\usepackage{adjustbox}
\usepackage{bbding}
\usepackage{pifont}
\usepackage{wasysym}
\usepackage{amssymb}
\usepackage[breaklinks=true,colorlinks,bookmarks=false]{hyperref}

\makeatletter
  \newcommand\figcaption{\def\@captype{figure}\caption}
  \newcommand\tabcaption{\def\@captype{table}\caption}
\makeatother
\newcommand\blfootnote[1]{%
  \begingroup
  \renewcommand\thefootnote{}\footnote{#1}%
  \addtocounter{footnote}{-1}%
  \endgroup
}
\usepackage[accsupp]{axessibility}  

\begin{document}

\title{The Devil is in the Few Shots: Iterative Visual Knowledge Completion for Few-shot Learning}

\titlerunning{Abbreviated paper title}

\author{Yaohui Li$^{\ast}$\inst{1,2} \and Qifeng Zhou$^{\ast}$\inst{1,2} \and Haoxing Chen\inst{3} \and Jianbing Zhang$^{\dagger}$\inst{1,2} \and \\ Xinyu Dai\inst{1,2} \and Hao Zhou\inst{4} }

\authorrunning{F.~Author et al.}

\institute{National Key Laboratory for Novel Software Technology, Nanjing University, China \and School of Artificial Intelligence, Nanjing University, China \and Ant Group
\and Institute of AI Industry Research (AIR), Tsinghua University, China
\email{\{yaohuili,zhouqf\}@smail.nju.edu.cn}}

\maketitle

\begin{abstract}
Contrastive Language-Image Pre-training (CLIP) has shown powerful zero-shot learning performance. Few-shot learning aims to further enhance the transfer capability of CLIP by giving few images in each class, aka “few shots”. Most existing methods either implicitly learn from the few shots by incorporating learnable prompts or adapters, or explicitly embed them in a cache model for inference. However, the narrow distribution of few shots often contains incomplete class information, leading to biased visual knowledge with high risk of misclassification. To tackle this problem, recent methods propose to supplement visual knowledge by generative models or extra databases, which can be costly and time-consuming. In this paper, we propose an Iterative Visual \textbf{K}nowledge \textbf{C}omp\textbf{L}etion (\textbf{KCL}) method to complement visual knowledge by properly taking advantages of unlabeled samples without access to any auxiliary or synthetic data. Specifically, KCL first measures the similarities between unlabeled samples and each category. Then, the samples with top confidence to each category is selected and collected by a designed confidence criterion. Finally, the collected samples are treated as labeled ones and added to few shots to jointly re-estimate the remaining unlabeled ones. The above procedures will be repeated for a certain number of iterations with more and more samples being collected until convergence, ensuring a progressive and robust knowledge completion process. Extensive experiments on 11 benchmark datasets demonstrate the effectiveness and efficiency of KCL as a plug-and-play module under both few-shot and zero-shot learning settings. Code is available at \href{https://github.com/Mark-Sky/KCL}{https://github.com/Mark-Sky/KCL} 
\end{abstract}
\blfootnote{$^{\ast}$ indicates equal contributions, $\dagger$ indicates corresponding author}
\section{Introduction}
\label{sec:intro}
Vision-language pre-training has significantly impacted the machine learning community, with various Vision-Language Models (VLMs) \cite{radford2021learning, li2022blip, jia2021scaling, li2023scaling, yu2022coca} demonstrating exceptional transfer performance across tasks such as classification \cite{khattak2023maple, chen2023sparse}, object detection \cite{wang2023learning, liu2023clip}, and segmentation \cite{wang2022cris, liu2023clip, zhou2023zegclip}. 
The most representative study, Contrastive Language-Image Pretraining (CLIP) \cite{radford2021learning}, is trained on a substantial dataset of image-text pairs using contrastive loss, have emerged as the standard for transferring knowledge to downstream tasks in computer vision. CLIP rebuilds the paradigm of zero-shot transfer, enabling classification on previous unseen categories according to similarities between image-text pairs. CLIP converts class labels into textual prompts, such as `\emph{a photo of a <class>}.', where \emph{<class>} represents the ground-truth textual label for each class. It then calculates similarities between all class prompts and the test image, ultimately selecting the class with the highest similarity score as the predicted label.
\begin{figure}[t]
  \centering
  \begin{subfigure}{0.55\linewidth}
    \includegraphics[width=0.93\linewidth]{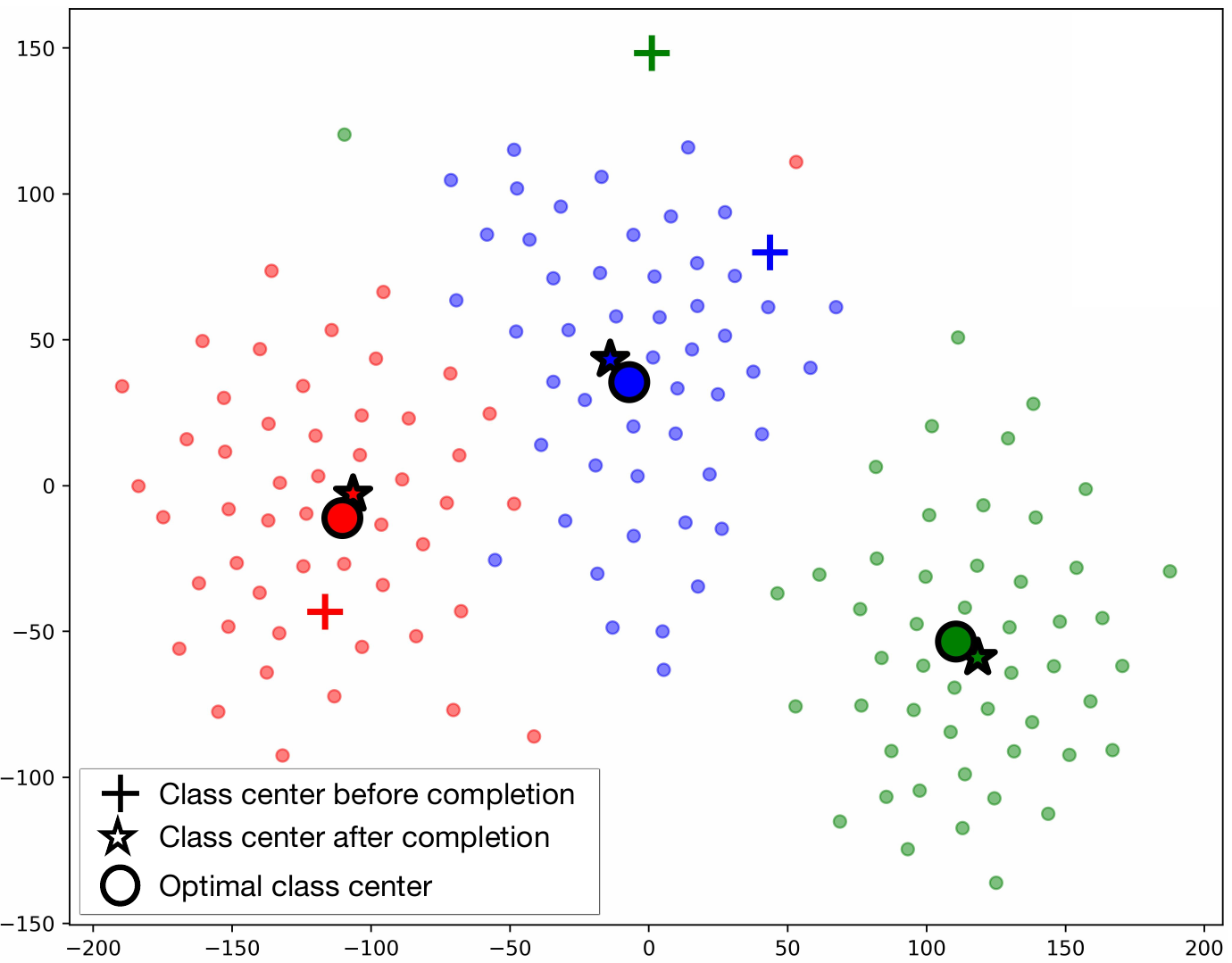}
    \caption{Feature visualization of 1-shot ImageNet \cite{deng2009imagenet}.}
    \label{figure:1-a}
  \end{subfigure}
  \begin{subfigure}{0.44\linewidth}
    \includegraphics[width=0.94\linewidth]{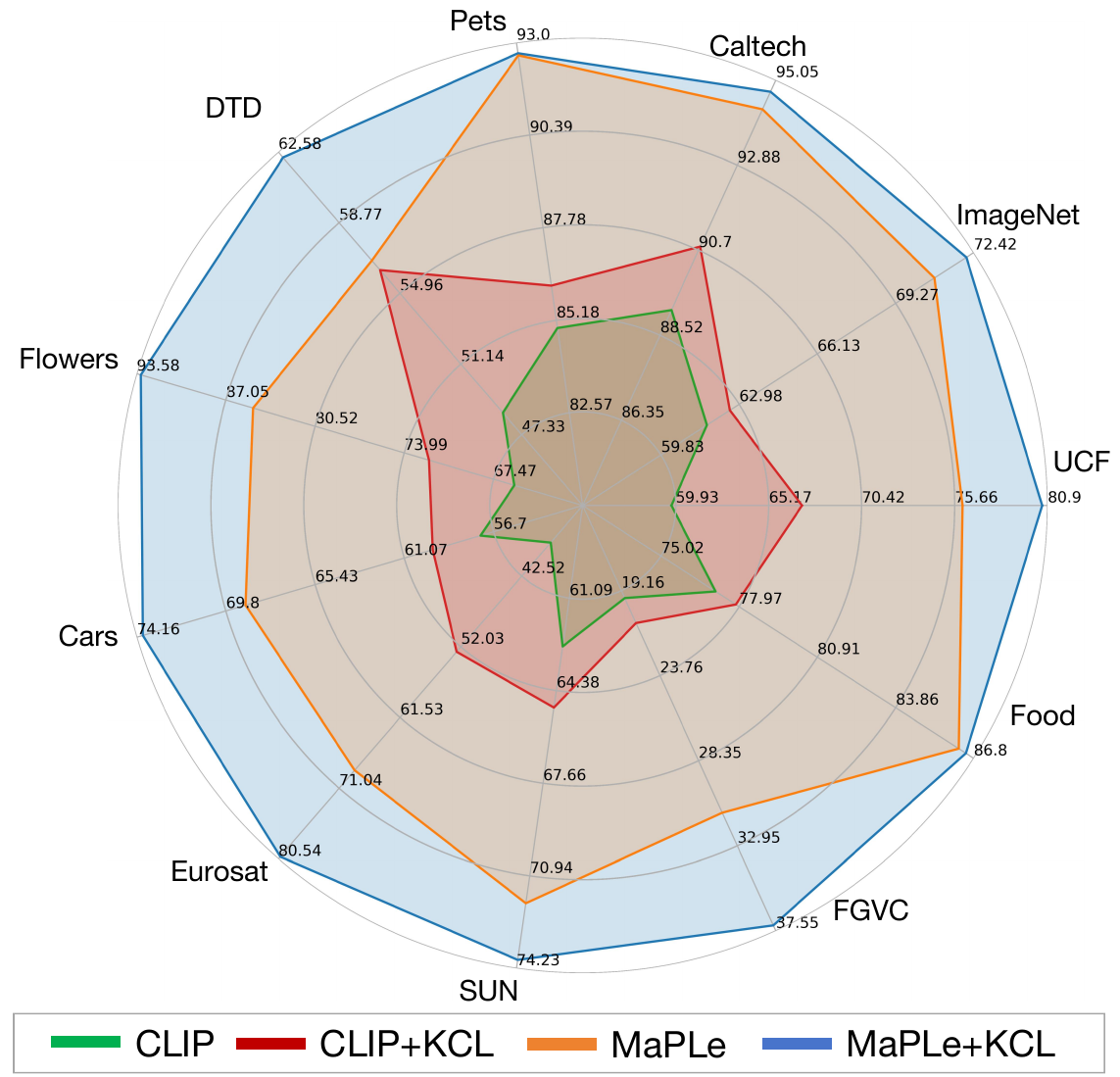}
    \caption{Performance comparison on 11 datasets.}
    \label{figure:1-b}
  \end{subfigure}
  \hfill
  \caption{Effectiveness demonstration of KCL. \textbf{(a)} t-SNE visualization of three categories in ImageNet \cite{deng2009imagenet} before and after iterative visual knowledge completion based on 1-shot MaPLe \cite{khattak2023maple}. \textbf{(b)} Classification results before and after using iterative visual knowledge completion on zero-shot CLIP \cite{radford2021learning} and 1-shot MaPLe \cite{khattak2023maple}.}
  \label{fig:1}
\end{figure}
Building upon the remarkable zero-shot transfer performance, recent studies have delved deeper and focus on enhancing CLIP within few-shot learning scenarios.
Under such settings, the model is tasked with further learning from a limited number of images, aka few shots, for each class to improve transfer capability towards novel tasks or domains. 
Existing methods mainly follow two paradigms to extract and utilize knowledge from few shots: Implicit modeling and explicit modeling. 
Specifically, implicit modeling-based methods mainly apply small number of learnable parameters to learn from few shots efficiently. 
Differently, explicit modeling-based methods do not need learnable parameters but follow the idea of cache models \cite{zhang2022tip}, which store the features and labels of few shots, and then conduct inference by similarity-based retrieval.

Despite the significant improvements made by existing few-shot learning approaches, they mainly focus on drawing more knowledge from the given few shots. 
However, the fundamental problem caused by data scarcity is rarely taken seriously, namely, the narrow-size distribution of few shots only contains incomplete and biased category information. 
As is shown in Fig. \ref{figure:1-a}, we can find that the given few shots are very likely to be distributed near classification boundaries and far away from the optimal class centers, especially in 1-shot scenario. 
In this case, the biased knowledge extracted from few shots can easily cause confusions with other nearby categories, leading to misclassification. Obviously, this situation also exists in zero-shot settings where no visual data is available \cite{udandarao2023sus}, which can be regarded as a special case of few-shot learning.

Recently, C2A \cite{roy2023Cap2Aug} and SuS-X \cite{udandarao2023sus} propose using generative models (e.g. Stable Diffusion \cite{rombach2022high, chen2024diffute}) and searching in large-scale external databases (e.g. LAION-5B \cite{schuhmann2022laion}) to alleviate visual data scarcity. However, generative models always involve large amount of computations, which can greatly affects efficiency. Similarly, auxiliary databases require additional storage space and searching through massive data is time-consuming as well. Besides, they both have ability bottlenecks and are prone to failure when facing previously unseen concepts \cite{udandarao2023sus}. Therefore, we ask the following question: 

\emph{Is it possible to complement the biases arising from data scarcity while also maintaining an efficient and concise process, without relying on generative models or auxiliary databases that may incur heavy costs and limited coverage?}

To achieve this goal, we propose an iterative visual knowledge completion method, termed as KCL, which can alleviate the bias caused by data scarcity during inference time without additional training. 
In contrast to existing methods that involve high-computation operations such as generation or searching in external databases, KCL offers an efficient solution to visual knowledge completion by incorporating high-confidence unlabeled samples with a designed confidence rule.
Specifically, KCL first calculates the similarity matrix between each test samples and all categories by integrating the similarity matrices from both textual and visual feature spaces.
Then, the most confident samples belonging to each class are selected as candidate samples with a design confidence criterion. 
Finally, the candidate samples are merged into few shots as supplements, refining potential class centers. The complemented knowledge will be used to re-estimate the remaining unlabeled data.
We repeat the above procedures for a certain number of iterations with growing number of samples being merged into few shots, guaranteeing a gradual and robust visual knowledge complementing process. 
As is shown in Fig. \ref{figure:1-a}, the potential class centers after knowledge completion are much closer to the optimal ones. 

We further extend KCL to zero-shot settings by modifying the first similarity matrix to only take textual knowledge from original CLIP, which can be viewed as a special case of few-shot learning.
Since KCL conducts adaptation at inference time, it can function as a plug-and-play module, seamlessly integrating with existing CLIP transfer methods without additional modifications. 
Therefore, we choose mainstream few-shot transfer methods from different types as few-shot baselines and original CLIP as zero-shot baseline. With the assistance of KCL, all these methods achieve remarkable improvements, which will be presented in Sec. \ref{exp}.
As is shown in Fig. \ref{figure:1-b}, we can see that KCL improves the performance of original methods (zero-shot CLIP \cite{radford2021learning} and MaPLe \cite{khattak2023maple}) by large margins, demonstrating its effectiveness in both few-shot and zero-shot settings.

Overall, our contributions are summarized as follows:

\begin{itemize}
\item[1.]  We figure out the fundamental factor: narrow-size distribution of few shots results in biased visual knowledge, that limits the transfer capability of CLIP.

\item[2.] We propose KCL, an iterative visual knowledge completion method which complements the biased visual knowledge without access to auxiliary and synthetic data. It takes advantages of a designed confidence criterion based on mutual nearest neighbors theory to mine useful knowledge from high-confidence test samples to supplement few shots iteratively.

\item[3.] We further extend KCL to zero-shot settings where no labeled data is available and enhance zero-shot CLIP by incorporating high-confidence visual knowledge from unlabeled data.

\item[4.] We conduct extensive experiments on 11 benchmark datasets, which demonstrate the effectiveness of KCL as a training-free plug-and-play module for both zero-shot and few-shot learning tasks.
\end{itemize}

\section{Related Works}
\label{sec:rw}
Vision-Language Models (VLMs) \cite{radford2021learning, li2022blip, jia2021scaling, li2023scaling, yu2022coca, yang2022vision, huang2023nlip, geng2023hiclip, seth2023dear, fan2024improving} have emerged as a novel paradigm in recent years. CLIP \cite{radford2021learning}, a groundbreaking study, trained on 400 million web-scale image-text pairs, showcasing remarkable performance on downstream tasks. In this work, we propose to enhance the transfer capability of CLIP, thus we will introduce existing transfer methods for CLIP with limited data \cite{zhou2022learning, zhou2022conditional, gao2023clip, chen2022plot, khattak2023maple, yao2023visual, wang2024hard, zhang2022tip, zhu2023not, pratt2023does}. Generally, we divide existing methods into two types depending on whether knowledge completion is adopted.

\subsection{Methods without Knowledge Completion}
Existing methods mainly focus on knowledge extraction other than completion.

\textbf{Implicit Modeling Methods.}
Implicit modeling methods mainly apply small number of learnable parameters to perform fine tuning, such as prompt tuning and feature adaptation. 
Unlike traditional methods that rely solely on pre-defined prompts, CoOp \cite{zhou2022learning} leverages back-propagation on few-shot datasets to dynamically generate text prompts tailored to specific tasks. 
On top of CoOp, CoCoOp \cite{zhou2022conditional} leverages visual features to regularize the learning process of text prompts, enabling model to capture both linguistic and visual cues simultaneously. 
Beyond textual prompt tuning, MaPLe \cite{khattak2023maple} propose a multi-modal prompt tuning strategy for both vision and language branches to improve alignment between visual and textual representations.
Meanwhile, CLIP-Adapter \cite{gao2023clip} involves learning a visual and a textual adapter, which are modules designed to refine the representations of the original CLIP specifically. 

\textbf{Explicit Modeling Methods.}
Explicit modeling methods follow the paradigm of metric learning and cache modeling.
Tip-Adapter \cite{zhang2022tip} first-time introduces the idea of explicit modeling for CLIP by directly constructing an adapter with a key-value cache model, which takes training-free advantage of zero-shot CLIP while also achieves remarkable performance compared to few-shot transfer methods.
On top of Tip-Adapter, APE \cite{zhu2023not} proposes to use a channel-wise feature selection strategy, which maintains the class-relevant features and wipes out class-irrelevant features. Recently, \cite{wang2024hard} proposes using Gaussian discriminant analysis to estimate parameters for constructing cache models.
\subsection{Methods with Knowledge Completion}
In general, existing transfer methods for CLIP mainly focus on how to extract more useful knowledge from few shots. However, the narrow-size distribution of few shots always contains incomplete class information, which may lead to naturally biased visual knowledge. CuPL \cite{pratt2023does} proposes to augment textual prompts by external knowledge bases (e.g. GPT-3) to enhance the understanding towards categories with limited visual data. Recent works SuS-X \cite{udandarao2023sus} and C2A \cite{roy2023Cap2Aug} propose using text-image and image-image generation \cite{rombach2022high} or auxiliary database LAION-5B \cite{schuhmann2022laion} to tackle visual data scarcity, which may involve large amount of computation and storage space. Differently, in this work, we propose to alleviate data scarcity by incorporating high-confidence test samples iteratively without using auxiliary or synthetic data.

\section{Methodology}
\label{sec:pm}
To better illustrate our iterative visual knowledge completion method, we first briefly revisit zero-shot CLIP \cite{radford2021learning} and few-shot Tip-Adapter \cite{zhang2022tip} methods.

\textbf{Zero-shot CLIP.} Given a classification task with $C$ categories, CLIP transforms the category names into textual prompts. The encoded prompt vectors can be interpreted as a weight matrix for the classifier, termed as $\bm{W} \in \mathbb{R}^{C \times d}$. Then, CLIP's visual encoder $E_v$ is applied to extract features from $N$ test samples $\{x_1, x_2, \cdots, x_N\}$:
\begin{equation}
    f_i = E_v (x_i), i\in [1, N], f_i\in \mathbb{R}^{1 \times d}.
\end{equation}
Finally, the cosine similarity matrix $\mathcal{A}_{clip}$ between $N$ test samples and $C$ categories can be computed by:
\begin{equation}
\label{clip}
    \bm{A}_{clip} = \bm{f W} \in \mathbb{R}^{N \times C},
\end{equation}
where $\bm f$ is the concatenation of $N$ image features $\{f_1, f_2, \cdots, f_N\}$.

\textbf{Tip-Adapter.} Under the setting of few-shot learning, zero-shot CLIP is supplemented with $M$ labeled samples for each category with one-hot label matrix $\bm{l} \in \mathbb{R}^{CM \times C}$. We also extract features of these $CM$ labeled samples $\{y_1, y_2, \cdots, y_{CM}\}$:
\begin{equation}
    F_i = E_v (y_i), i\in [1, CM], F_i\in \mathbb{R}^{1 \times d}.
\end{equation}
Then the visual similarity matrix between the concatenated visual features $\bm f$ and $\bm F$ is computed by $\text{exp}(-\beta(1-\bm{f}\bm{F}^{\top})$, where $\beta$ is a dataset-specific hyper-parameter.
Finally, the overall similarity matrix $\mathcal{A}_{tip}$ is given by:
\begin{equation}
\label{tip}
    \bm{A}_{tip} = \alpha \text{exp}{(-\beta(1-\bm{fF}^{\top}))} \bm{l} + \bm{f W} \in \mathbb{R}^{N \times C},
\end{equation}
where $\alpha$ is a hyper-parameter controlling the ratio between visual and textual similarity matrices.

\textbf{Methodology of KCL.} Motivated by C2A \cite{roy2023Cap2Aug} and SuS-X \cite{udandarao2023sus} which augment visual data by utilizing generative diffusion models or large-scale external databases. In this work, we aim to complement the biased visual knowledge of few shots by taking advantages of unlabeled samples without accessing auxiliary or synthetic data. 

Following \cite{roy2023Cap2Aug} and \cite{udandarao2023sus}, we also build our KCL module upon Tip-Adapter. Eq. \ref{tip} can be viewed as two parts, respectively standing for the image-to-image and image-to-text similarities between $N$ test features $\bm{f}=\{f_1, f_2, \cdots, f_N\}$ and $C$ categories $\bm{S}=\{S_1, S_2, \cdots, S_C\}$. As is discussed before, the knowledge in few-shot visual features $\bm F$ is biased due to data scarcity and we propose to complement $\bm F$ by using high-confidence features in $\bm f$. Therefore, the confidence criterion used to distinguish the most confident samples becomes a key issue. 

In this work, we design a confidence criterion based on mutual nearest neighbors theory \cite{gowda1979condensed, chen2022mutual}, which means only when a test sample and the category are nearest neighbors of each other, will it be regarded as a high-confidence sample and selected. Specifically, given the $i$-th category $S_i$ with label $l_i$, we can define its mutual nearest neighbors (MNNs) in $\bm f$ as:
\begin{equation}
\label{eq5}
    \mathcal{M}_{K_1,K_2}(S_i) = \{f_j \in \bm{f}|f_j \in \mathcal{N}_{K_1}^{\bm f}(S_i) \wedge S_i \in \mathcal{N}_{K_2}^{\bm S}(f_j)\},
\end{equation}
where $\mathcal{N}_{K}^{\bm B}(a)$ represents a subset of $\bm B$ consisting of the $K$ nearest neighbors of $a$ in $\bm B$. Due to the uniqueness of classification, we fix $K_2=1$ under all situations and will further discuss the influence of $K_1$ towards effectiveness and efficiency in Sec. \ref{ablation} and Sec. \ref{dis}. After seeking out MNNs for all categories, we get $\{\mathcal{M}_{K_1,K_2}(S_i)\}_{i=1}^{C} \in \bm{f}$, then we label them with corresponding categories and collect them into a knowledge completion set $\bm{D}$ with label set $\bm l_D$, completing one round of knowledge completion. The knowledge completion set $\bm D$ will be utilized to jointly re-estimate the similarity matrix between the remaining test data $\bm f'=\bm{f}-\bm{D}\in \mathbb{R}^{N' \times d}$ and $C$ categories ($N'$ denotes the number of remaining samples in $\bm f'$).
\begin{equation}
\label{eq6}
    \bm{A}_{kcl}^{fs} = \lambda \text{exp}{(-\mu(1-\bm{f'D}^{\top}))} \bm{l_D} + \alpha \text{exp}{(-\beta(1-\bm{f'F}^{\top}))} \bm{l} + \bm{f' W} \in \mathbb{R}^{N' \times C}.
\end{equation}
Note that $\lambda$ and $\mu$ are hyper-parameters for $\bm D$, which will be discussed in Sec. \ref{Exp-r}. Besides, $\alpha$ and $\beta$ are hyper-parameters in Eq. \ref{tip}, which are fixed from beginning. In practice, we repeat the above procedures for a certain number of iterations to gradually increases the number of samples taken from $\bm f$ into $\bm D$, ensuring a robust and progressive knowledge completion process. The specific number of iterations or the number of unlabeled samples required for visual knowledge completion will be discussed in Sec. \ref{ablation} and Sec. \ref{dis}.

We further extend KCL to zero-shot learning scenarios where $\bm{F}=\emptyset$ and the first-step similarity matrix is given by Eq. \ref{clip}. Then the similarity matrix after confidence selection is modified to:
\begin{equation}
\label{eq7}
    \bm{A}_{kcl}^{zs} = \lambda \text{exp}{(-\mu(1-\bm{f'D}^{\top}))} \bm{l_D} + \bm{f' W} \in \mathbb{R}^{N' \times C}.
\end{equation}
Note that we keep the same hyper-parameter searching mechanism as few-shot learning and also repeat a certain number of iterations until convergence. 

Note that, when reaching convergence, either few-shot or zero-shot classification results will be given by $\bm l_D$ and the last-round similarity matrix based on the nearest neighbor theory.

\section{Experiments}
\label{exp}

\subsection{Experimental Settings}
\textbf{Datasets.}
In this work, following the common practice in CLIP transfer scenario, we adopt 11 benchmark datasets spanning a wide range of object, scene and fine-grained categories for a comprehensive evaluation: ImageNet \cite{deng2009imagenet}, UCF \cite{soomro2012ucf101}, DTD \cite{cimpoi2014describing}, FGVC \cite{maji2013fine}, Flowers \cite{nilsback2008automated}, Cars \cite{krause20133d}, Eurosat \cite{helber2019eurosat}, SUN \cite{xiao2010sun}, Pets \cite{parkhi2012cats}, Caltech \cite{fei2004learning} and Food \cite{bossard2014food}. 

\textbf{Implementation details.}
For few-shot transfer, we follow \cite{zhou2022learning} and conduct experiments on 1, 2, 4, 8, 16 few-shot classification tasks across 11 datasets. Since KCL acts as a plug-and-play module, we follow the backbone settings in original papers of each method. Specifically, for experiments on zero-shot CLIP, CoOp, CLIP-Adapter, Tip-Adapter and APE, we choose ResNet-50 \cite{he2016deep} as backbone; while for experiments on MaPLe, we choose ViT-B/16 \cite{dosovitskiy2020image} as visual backbone structure. For textual prompts, we utilize CuPL \cite{pratt2023does} for all experimental results in this work for fair comparisons between different approaches. Besides, we follow the training details of CoOp \cite{zhou2022conditional}, CLIP-Adapter \cite{gao2023clip} and MaPLe \cite{khattak2023maple} in original papers. The hyper-parameter settings of this work are as follows: for $\alpha$ and $\beta$, we follow \cite{zhang2022tip} and search on validation set at the first round, which will be fixed in following procedures; while for newly introduced $\lambda$ and $\mu$, we search on validation set for the best numerical values in each iteration. All experiments in this work are conducted on a single NVIDIA Tesla V100-32G GPU.

\subsection{Baseline Methods}
Since KCL can function as a plug-and-play module, we are able to embed it to any type of approaches. Therefore, to demonstrate its effectiveness, we embed KCL to representative methods for CLIP-based few-shot learning of different types as well as zero-shot CLIP.

\textbf{Few-shot learning.}
\textbf{CoOp} \cite{zhou2022learning} is the most representative method in the field of textual prompt tuning for CLIP transfer, which proposes to replace fixed prompts with learnable text tokens. Note that we initialize the learnable tokens of CoOp with CuPL \cite{pratt2023does} in this work. \textbf{CLIP-Adapter} \cite{gao2023clip} is the representative method in the field of feature adapters for CLIP transfer, which adds MLPs to the two branches of CLIP and adapts by gradient decent on these two MLPs. We follow the settings of the original paper for CLIP-Adapter in this paper. \textbf{Tip-Adapter} \cite{zhang2022tip} is a representative approach in the field of cache model-based CLIP transfer, which explicitly encodes few-shot data into a cache model and propagate knowledge to test samples. We also follow the original settings of Tip-Adapter to construct the key-value cache model. \textbf{MaPLe} \cite{khattak2023maple} is a representative method of multi-modal prompt tuning, which combines learnable textual and visual prompts to learn better visual and textual representations for few-shot transfer. We also include \textbf{APE} \cite{zhu2023not} in supplementary materials, which extends Tip-Adapter with channel-wise feature refinement. We attach KCL to above methods and improvements are shown in Table \ref{tab1}. 

\textbf{Zero-shot learning.}
\textbf{Zero-shot CLIP } \cite{radford2021learning}: we use prompt ensembling strategy
with seven prompt templates following \cite{udandarao2023sus}. \textbf{CALIP} \cite{guo2023calip} proposes using a cross attention mechanism to obtain more discriminative representations. \textbf{CuPL} \cite{pratt2023does} augments textual prompts of original CLIP by consulting GPT-3. \textbf{SuS-X} \cite{udandarao2023sus} proposes to complement visual knowledge of zero-shot CLIP by diffusion-based image generation or retrieval in large-scale databases. Experimental results under zero-shot settings are shown in Table \ref{tab2}. 

\begin{table}[t]
\caption{Experimental results (\%) of few-shot learning on 11 benchmark datasets. The overall results including 8-shot and 16-shot are shown in supplementary materials.}
\resizebox{\textwidth}{!}{
\begin{tabular}{lccccccccccc|c}
\toprule[1pt]
  \textbf{Method}& \rotatebox{60}{\textbf{ImageNet}} & \rotatebox{60}{\textbf{UCF}} & \rotatebox{60}{\textbf{DTD}} & \rotatebox{60}{\textbf{FGVC}} & \rotatebox{60}{\textbf{Flowers}} & \rotatebox{60}{\textbf{Cars}} & \rotatebox{60}{\textbf{Eurosat}} & \rotatebox{60}{\textbf{SUN}} & \rotatebox{60}{\textbf{Pets}} & \rotatebox{60}{\textbf{Caltech}} & \rotatebox{60}{\textbf{Food}} & \rotatebox{60}{\textbf{Average}} \\ \midrule
\multicolumn{13}{c}{\textbf{1-shot}}\\ \midrule
\textbf{CoOp} \cite{zhou2022learning}        & 55.70                              & 64.92                        & 53.25                        & 21.18                         & 81.81                            & 53.89                         & 62.56                            & 60.68                           & 81.60                          & 90.06                            & 73.42                         & 63.55                            \\
$\bm\Delta \textbf{+KCL}$        & {\color[HTML]{009901} +2.84}      & {\color[HTML]{009901} +4.73} & {\color[HTML]{009901} +4.73} & {\color[HTML]{009901} +0.18}  & {\color[HTML]{009901} +4.22}     & {\color[HTML]{009901} +3.40}   & {\color[HTML]{009901} +6.75}     & {\color[HTML]{009901} +2.77}    & {\color[HTML]{009901} +3.11}  & {\color[HTML]{009901} +0.49}     & {\color[HTML]{009901} +2.76}  & {\color[HTML]{009901} +3.27}     \\ \midrule
\textbf{CLIP-Adapter} \cite{gao2023clip}       & 61.94                             & 64.58                        & 51.60                         & 21.27                         & 72.03                            & 58.77                         & 61.23                            & 63.83                           & 85.61                         & 90.10                             & 76.80                          & 64.34                            \\
$\bm\Delta \textbf{+KCL}$       & {\color[HTML]{009901} +1.11}      & {\color[HTML]{009901} +6.90}  & {\color[HTML]{009901} +5.97} & {\color[HTML]{009901} +1.47}  & {\color[HTML]{009901} +10.63}    & {\color[HTML]{009901} +1.67}  & {\color[HTML]{009901} +0.99}     & {\color[HTML]{009901} +1.83}    & {\color[HTML]{009901} +1.17}  & {\color[HTML]{009901} +1.06}     & {\color[HTML]{009901} +0.69}  & {\color[HTML]{009901} +3.04}     \\ \midrule
\textbf{Tip-Adapter} \cite{zhang2022tip}       & 61.83                             & 63.86                        & 51.12                        & 20.94                         & 72.63                            & 58.60                          & 57.68                            & 54.08                           & 85.25                         & 89.25                            & 77.57                         & 63.89                            \\
$\bm\Delta \textbf{+KCL}$      & {\color[HTML]{009901} +0.71}      & {\color[HTML]{009901} +6.24} & {\color[HTML]{009901} +5.44} & {\color[HTML]{009901} +0.63}  & {\color[HTML]{009901} +9.79}     & {\color[HTML]{009901} +2.74}  & {\color[HTML]{009901} +7.05}     & {\color[HTML]{009901} +1.50}     & {\color[HTML]{009901} +1.20}   & {\color[HTML]{009901} +1.34}     & {\color[HTML]{009901} +0.61}  & {\color[HTML]{009901} +3.38}     \\ \midrule
\textbf{MaPLe} \cite{khattak2023maple}      & 69.67                             & 69.94                        & 41.31                        & 27.18                         & 77.30                             & 66.99                         & 60.75                            & 70.05                           & 91.55                         & 93.06                            & 84.12                         & 68.35                            \\
$\bm\Delta \textbf{+KCL}$        & {\color[HTML]{009901} +1.11}      & {\color[HTML]{009901} +3.57} & {\color[HTML]{009901} +6.56} & {\color[HTML]{009901} +2.73}  & {\color[HTML]{009901} +12.87}    & {\color[HTML]{009901} +3.92}  & {\color[HTML]{009901} +6.78}     & {\color[HTML]{009901} +1.29}    & {\color[HTML]{009901} +0.27}  & {\color[HTML]{009901} +0.37}     & {\color[HTML]{009901} +0.22}  & {\color[HTML]{009901} +3.60}     \\ \midrule
\multicolumn{13}{c}{\textbf{2-shot}}\\ \midrule
\textbf{CoOp} \cite{zhou2022learning}& 57.36                             & 68.44                        & 55.85                        & 25.11                         & 84.94                            & 60.17                         & 65.49                            & 62.03                           & 84.19                         & 89.49                            & 73.58                         & 66.05                            \\
$\bm\Delta \textbf{+KCL}$       & {\color[HTML]{009901} +1.15}      & {\color[HTML]{009901} +4.97} & {\color[HTML]{009901} +0.53} & {\color[HTML]{CB0000} -0.18}  & {\color[HTML]{009901} +3.61}     & {\color[HTML]{009901} +1.03}  & {\color[HTML]{009901} +7.52}     & {\color[HTML]{009901} +3.17}    & {\color[HTML]{009901} +1.36}  & {\color[HTML]{009901} +0.17}     & {\color[HTML]{009901} +2.28}  & {\color[HTML]{009901} +2.32}     \\ \midrule
\textbf{CLIP-Adapter} \cite{gao2023clip}      & 61.84                             & 66.88                        & 56.97                        & 21.87                         & 73.12                            & 58.92                         & 62.23                            & 64.11                           & 87.33                         & 90.83                            & 77.34                         & 65.58                            \\
$\bm\Delta \textbf{+KCL}$        & {\color[HTML]{009901} +1.00}      & {\color[HTML]{009901} +2.93} & {\color[HTML]{009901} +3.49} & {\color[HTML]{009901} +0.84}  & {\color[HTML]{009901} +3.25}     & {\color[HTML]{009901} +1.12}  & {\color[HTML]{009901} +5.89}     & {\color[HTML]{009901} +1.21}    & {\color[HTML]{009901} +0.62}  & {\color[HTML]{009901} +0.08}     & {\color[HTML]{009901} +0.54}  & {\color[HTML]{009901} +1.90}     \\ \midrule
\textbf{Tip-Adapter} \cite{zhang2022tip}       & 62.01                             & 66.56                        & 54.85                        & 23.10                         & 76.49                            & 60.87                         & 59.07                            & 65.49                           & 84.25                         & 89.49                            & 77.35                          & 65.32                            \\
$\bm\Delta \textbf{+KCL}$       & {\color[HTML]{009901} +0.84}      & {\color[HTML]{009901} +4.57} & {\color[HTML]{009901} +2.36} & {\color[HTML]{009901} +0.48}  & {\color[HTML]{009901} +8.12}     & {\color[HTML]{009901} +1.40}   & {\color[HTML]{009901} +9.10}      & {\color[HTML]{009901} +1.16}    & {\color[HTML]{009901} +2.56}  & {\color[HTML]{009901} +0.53}     & {\color[HTML]{009901} +0.45}  & {\color[HTML]{009901} +2.87}     \\ \midrule
\textbf{MaPLe} \cite{khattak2023maple}       & 70.43                             & 73.96                        & 53.66                        & 28.98                         & 79.54                            & 63.06                         & 57.89                            & 70.35                           & 91.09                         & 93.87                            & 86.33                         & 69.92                            \\
$\bm\Delta \textbf{+KCL}$        & {\color[HTML]{009901} +1.17}      & {\color[HTML]{009901} +3.91} & {\color[HTML]{009901} +4.79} & {\color[HTML]{009901} +5.22}  & {\color[HTML]{009901} +12.75}    & {\color[HTML]{009901} +5.10}   & {\color[HTML]{009901} +20.64}    & {\color[HTML]{009901} +2.43}    & {\color[HTML]{009901} +0.46}  & {\color[HTML]{009901} +0.25}     & {\color[HTML]{009901} +0.58}  & {\color[HTML]{009901} +5.20}     \\ \midrule
\multicolumn{13}{c}{\textbf{4-shot}}\\ \midrule
\textbf{CoOp} \cite{zhou2022learning}  & 58.86                             & 70.71                        & 61.64                        & 26.37                         & 89.77                            & 64.71                         & 72.70                             & 65.82                           & 87.24                         & 91.48                            & 74.91                         & 69.47                            \\
$\bm\Delta \textbf{+KCL}$       & {\color[HTML]{009901} +1.64}      & {\color[HTML]{009901} +4.28} & {\color[HTML]{009901} +0.18} & 0.00                             & {\color[HTML]{009901} +1.50}      & {\color[HTML]{009901} +0.19}  & {\color[HTML]{009901} +4.29}     & {\color[HTML]{009901} +0.95}    & {\color[HTML]{009901} +0.39}  & {\color[HTML]{CB0000} -0.04}     & {\color[HTML]{009901} +1.20}   & {\color[HTML]{009901} +1.32}     \\ \midrule
\textbf{CLIP-Adapter} \cite{gao2023clip}      & 62.24                             & 69.15                        & 57.98                        & 22.20                          & 82.01                            & 60.81                         & 73.27                            & 66.74                           & 87.52                         & 91.16                            & 77.82                         & 68.26                            \\
$\bm\Delta \textbf{+KCL}$        & {\color[HTML]{009901} +0.98}      & {\color[HTML]{009901} +2.94} & {\color[HTML]{CB0000} -0.12} & {\color[HTML]{009901} +1.71}  & {\color[HTML]{009901} +6.42}     & {\color[HTML]{009901} +2.57}  & {\color[HTML]{009901} +5.15}     & {\color[HTML]{009901} +1.16}    & {\color[HTML]{009901} +0.35}  & {\color[HTML]{009901} +0.73}     & {\color[HTML]{009901} +0.41}  & {\color[HTML]{009901} +2.02}     \\ \midrule
\textbf{Tip-Adapter} \cite{zhang2022tip}       & 62.16                             & 68.46                        & 59.22                        & 23.34                         & 82.34                             & 62.82                         & 69.27                            & 65.57                            & 84.79                          & 90.34                            & 77.25                         & 67.77                            \\
$\bm\Delta \textbf{+KCL}$        & {\color[HTML]{009901} +0.49}      & {\color[HTML]{009901} +3.18} & {\color[HTML]{009901} +1.60}  & {\color[HTML]{CB0000} -0.09}  & {\color[HTML]{009901} +4.26}     & {\color[HTML]{009901} +1.02}  & {\color[HTML]{009901} +2.74}     & {\color[HTML]{009901} +1.13}    & {\color[HTML]{009901} +1.61}  & {\color[HTML]{009901} +0.49}     & {\color[HTML]{009901} +0.49}  & {\color[HTML]{009901} +1.53}     \\ \midrule
\textbf{MaPLe} \cite{khattak2023maple}       & 70.85                             & 76.26                        & 58.51                        & 31.80                          & 84.49                            & 69.71                         & 62.40                             & 72.00                              & 92.94                         & 94.44                            & 86.27                         & 72.69                            \\
$\bm\Delta \textbf{+KCL}$        & {\color[HTML]{009901} +1.44}      & {\color[HTML]{009901} +7.14} & {\color[HTML]{009901} +5.44} & {\color[HTML]{009901} +4.11}  & {\color[HTML]{009901} +9.26}     & {\color[HTML]{009901} +4.44}  & {\color[HTML]{009901} +16.59}    & {\color[HTML]{009901} +2.08}    & {\color[HTML]{009901} +0.03}  & {\color[HTML]{009901} +1.02}     & {\color[HTML]{009901} +0.14}  & {\color[HTML]{009901} +4.69} \\
\bottomrule[1pt]
\end{tabular}
}
\label{tab1}
\end{table}

\subsection{Experimental Results}
\label{Exp-r}
\begin{figure}[t]
  \centering
  \begin{subfigure}{0.49\linewidth}
    \includegraphics[width=\linewidth]{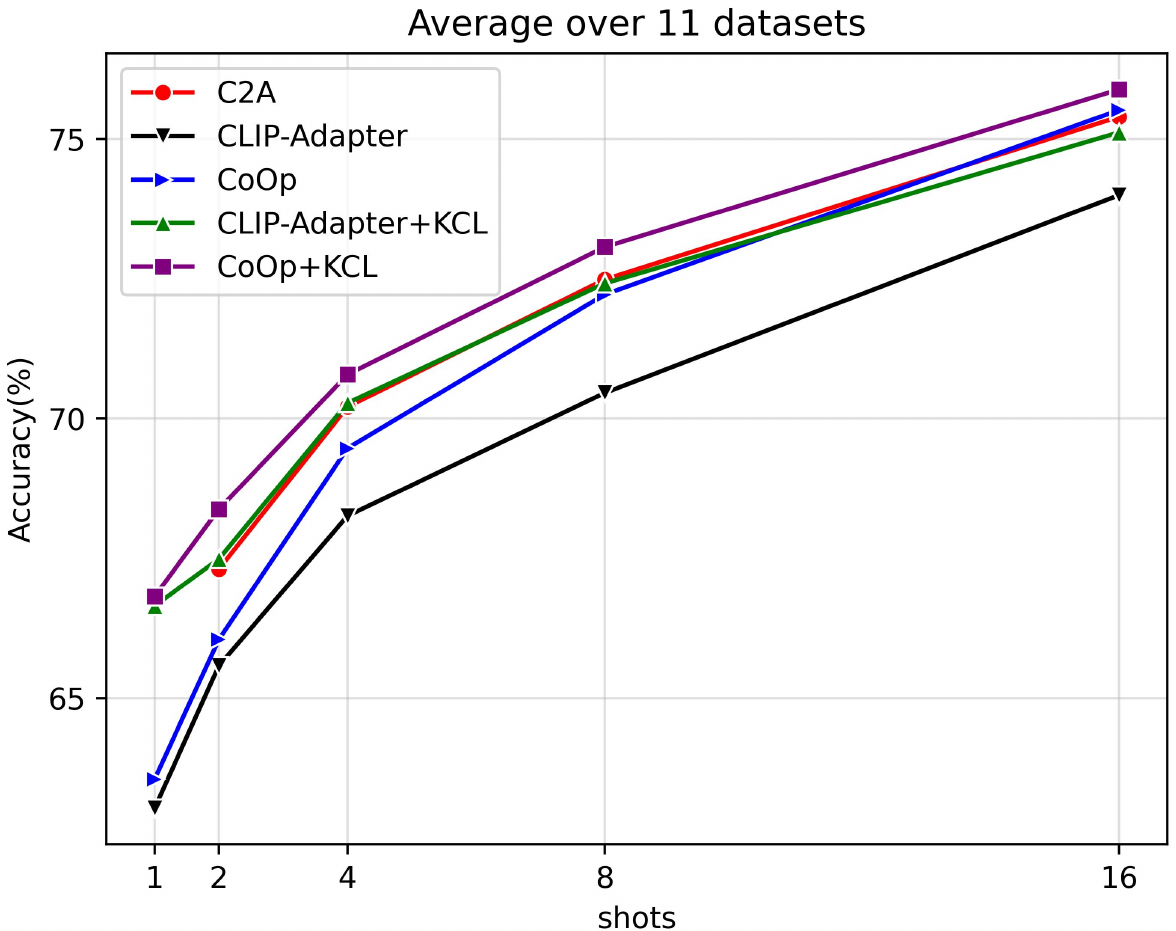}
    \caption{Analysis of knowledge completion capability.}
    \label{figure:2-a}
  \end{subfigure}
    \begin{subfigure}{0.49\linewidth}
    \includegraphics[width=0.99\linewidth]{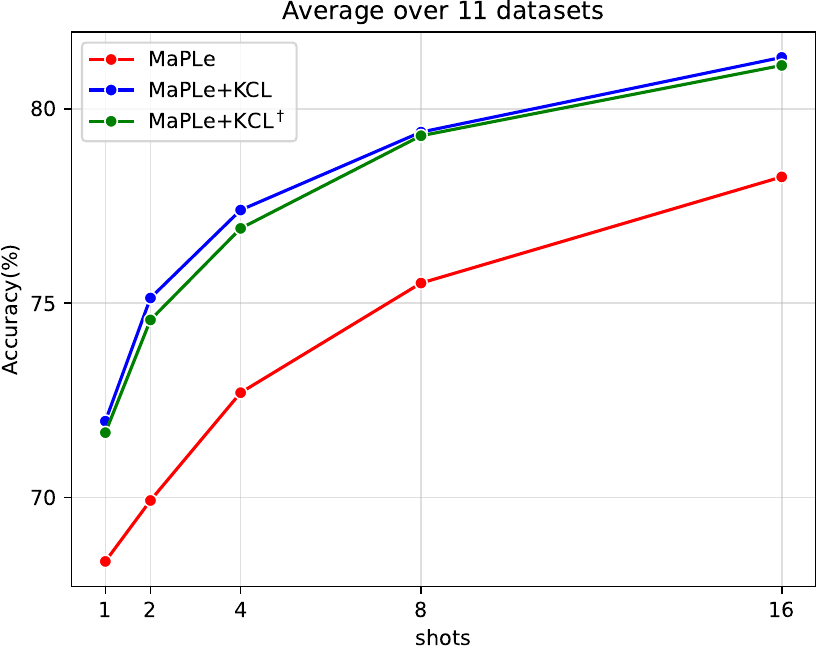}
    \caption{Analysis of hyper-parameter sensitivity.}
    \label{figure:2-b}
  \end{subfigure}
  \hfill
  \caption{Further analyses under few-shot settings. (a) The averaged performance on 11 datasets compared with C2A \cite{roy2023Cap2Aug}. (b) The averaged performance of hyper-parameter sensitivity analysis on 11 datasets based on MaPLe \cite{khattak2023maple}. Note that $^{\dagger}$ means hyper-parameters $\lambda$ and $\mu$ of KCL method are both fixed to 1.} 
  \label{fig:2}
\end{figure}

\noindent\textbf{Few-shot learning results.}
To estimate the few-shot learning capability of KCL as a plug-and-play module, we apply it to several kinds of CLIP-based few-shot learning methods, and a part of the experimental results are shown in Table \ref{tab1} (please refer tosupp for overall results). 
We can see that KCL significantly improves the average performances of existing methods on 11 datasets. Specifically, for 1-shot setting, KCL can achieve 3.04\%-3.60\% improvements; for 2-shot setting, KCL can gain 1.90\%-5.20\% improvements; for 4-shot setting, KCL achieves 1.32\%-4.69\% improvements over original methods; for 8-shot tasks, KCL can surpass the initial methods by 0.83\%-3.88\%; for 16-shot tasks, KCL achieves an average improvement of 1.11\%. 
Generally, we can find that KCL achieves relatively larger improvements on fewer-shot settings. 
This phenomenon is in line with our motivations for proposing KCL, which aims to complement biased visual knowledge caused by data scarcity. 
Naturally, the fewer the shots, the greater the knowledge bias, and thus the larger the improvements made by KCL.
Furthermore, we notice that KCL can achieve larger improvements when embedded to relatively stronger base models. This is because stronger base method can better distinguish unlabeled samples with high confidence.

Besides, we also compare KCL with existing visual knowledge completion methods C2A \cite{roy2023Cap2Aug}, which takes advantages of stable diffusion to generate images for knowledge completion. Since C2A method complements visual knowledge based on captions made by BLIP \cite{li2022blip}, Tip-Adater-F \cite{zhang2022tip} model and feature adapters, we compare it with KCL modules embedded in CoOp and CLIP-Adapter. As is shown in Fig. \ref{figure:2-a}, we can find that KCL module based on single CLIP-Adapter can achieve competitive performance against C2A. Besides, KCL module embedded in CoOp can surpass C2A by almost 1\% improvements under all settings. This undoubtedly proves the superiority of KCL which does not require additional generative models.

Furthermore, we also conduct hyper-parameter sensitivity analysis on 11 datasets based on MaPLe. As is shown in Fig. \ref{figure:2-b}, we can find that hyper-parameter searching has a small impact on final classification performance, which reveals the robustness and simplicity of the KCL as a plug-and-play module.

\begin{table}[t]
\caption{Experimental results (\%) of zero-shot learning on 11 benchmark datasets. We present the results reported in \cite{udandarao2023sus}.}
\resizebox{\textwidth}{!}{
\begin{tabular}{lccccccccccc|c}
\toprule[1pt] 
\textbf{Method}& \rotatebox{60}{\textbf{ImageNet}} & \rotatebox{60}{\textbf{UCF}} & \rotatebox{60}{\textbf{DTD}} & \rotatebox{60}{\textbf{FGVC}} & \rotatebox{60}{\textbf{Flowers}} & \rotatebox{60}{\textbf{Cars}} & \rotatebox{60}{\textbf{Eurosat}} & \rotatebox{60}{\textbf{SUN}} & \rotatebox{60}{\textbf{Pets}} & \rotatebox{60}{\textbf{Caltech}} & \rotatebox{60}{\textbf{Food}} & \rotatebox{60}{\textbf{Average}} \\ \midrule

\textbf{CLIP} \cite{radford2021learning}            & 60.31& 55.56& 41.01& 16.71& 62.89& 56.33& 26.83& 59.11& 81.82& 85.92& 74.11&56.41 \\ 
\textbf{CALIP} \cite{guo2023calip}                  & 60.31& 55.61& 41.01& 16.71& 63.01& 56.32& 26.96& 59.10& 81.84& 86.20& 74.13&56.47 \\ 
\textbf{CuPL} \cite{pratt2023does}                  & 61.45& 58.97& 48.64& 19.59& 65.44& 57.28& 38.38& 62.55& 84.84& 89.29& 76.94&60.30 \\ 
\textbf{CuPL-e} \cite{pratt2023does}                & 61.64& 61.45& 47.45& 19.26& 65.93& 57.23& 37.06& 62.99& 85.09& 89.41& 77.52&60.45 \\ 
\textbf{VisDesc} \cite{menon2022visual}             & 59.68& 58.47& 41.96& 16.26& 65.37& 54.76& 37.60& 59.84& 82.39& 88.11& 76.80&58.30 \\ 
\textbf{SuS-X}$\bm _{SD}$ \cite{udandarao2023sus}   & 61.84& 61.54& 50.59& 19.47& 67.72& 57.27& 45.57& 62.95& 85.34& 89.53& 77.58&61.72 \\ 
\textbf{SuS-X}$\bm _{LC}$ \cite{udandarao2023sus}   & 61.79& 61.43& 49.47& 20.34& 67.60& 57.06& 37.16& 62.94& 85.22& 89.61& 77.58&60.93 \\ \midrule
\textbf{KCL} (Ours)                            & \textbf{62.60}& \textbf{67.06}& \textbf{56.21}& \textbf{20.91}& \textbf{72.19}& \textbf{59.68}& \textbf{52.69}& \textbf{64.98}& \textbf{86.15}& \textbf{90.79}& \textbf{77.84}&\textbf{64.65} \\ 
$\bm \Delta$ vs \textbf{CLIP}                  & {\color[HTML]{009901} +2.29}&{\color[HTML]{009901} +11.50}&{\color[HTML]{009901} +15.20}& {\color[HTML]{009901} +4.20}&{\color[HTML]{009901} +9.30}&{\color[HTML]{009901}  +3.35}&{\color[HTML]{009901} +25.86}& {\color[HTML]{009901} +5.87}& {\color[HTML]{009901} +4.33}& {\color[HTML]{009901} +4.87}& {\color[HTML]{009901} +3.73}&{\color[HTML]{009901} +8.23} \\
$\bm \Delta$ vs \textbf{SuS-X}                & {\color[HTML]{009901} +0.76}& {\color[HTML]{009901} +5.52}& {\color[HTML]{009901} +5.62}& {\color[HTML]{009901} +0.57}& {\color[HTML]{009901} +4.47}& {\color[HTML]{009901} +2.41}& {\color[HTML]{009901} +7.12}& {\color[HTML]{009901} +2.03}& {\color[HTML]{009901} +0.81}& {\color[HTML]{009901} +1.18}& {\color[HTML]{009901} +0.26}&{\color[HTML]{009901} +2.82} \\

\bottomrule[1pt]
\end{tabular}
}
\label{tab2}
\end{table}
\textbf{Zero-shot learning results.}
To evaluate KCL on zero-shot learning tasks, we also conduct experiments on 11 benchmark datasets. As is shown in Table \ref{tab2}, we can observe that KCL achieves state-of-the-art performance on 11 benchmark datasets. Specifically, when compared with zero-shot CLIP, KCL gains 2.29\%-25.86\% improvements and 8.23\% improvements in average, demonstrating the effectiveness of iterative visual knowledge completion for CLIP. Besides, when compared with SuS-X \cite{udandarao2023sus} that complements visual knowledge with generative models and large-scale auxiliary databases, KCL also achieves 2.82\% improvements in average. This demonstrates the superiority of KCL against SuS-X, which can replenish visual knowledge by utilizing unlabeled samples without accessing auxiliary or synthetic data.

\subsection{Ablation Study}
\label{ablation}
In this section, we dive into the details of KCL method and analyze different choices of $K_1$ values (hereafter denoted as $K$) in Eq. \ref{eq5}, the number of unlabeled samples required for knowledge completion, the effectiveness of our confidence criterion and modalities used to calculate similarity matrix.

\textbf{The choice of $\bm K$ values.}
\begin{figure}[t]
  \centering
  \begin{subfigure}{0.32\linewidth}
    \includegraphics[width=\linewidth]{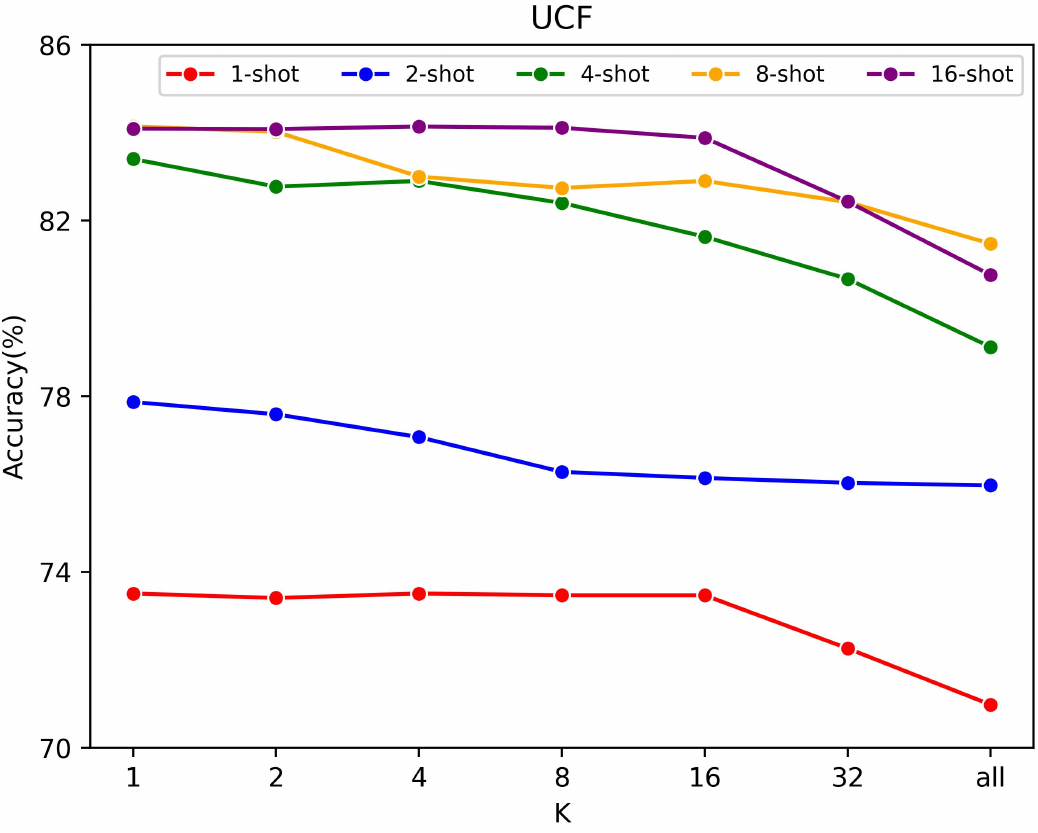}
    \caption{UCF}
    \label{fig:3-b}
  \end{subfigure}
   \begin{subfigure}{0.32\linewidth}
    \includegraphics[width=\linewidth]{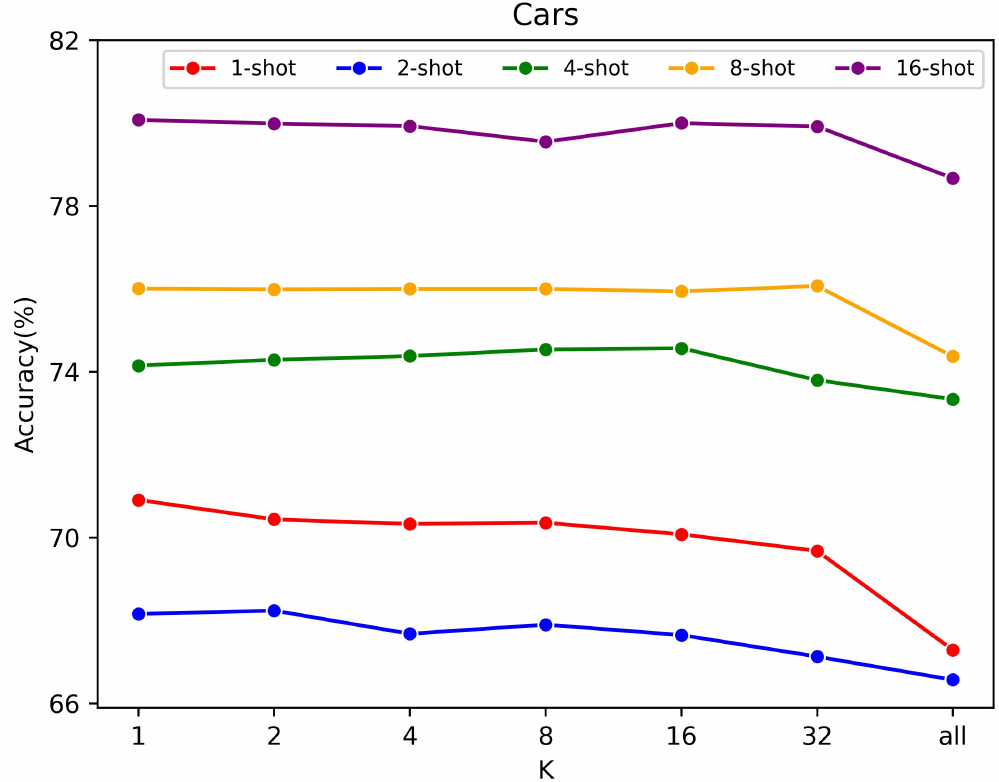}
    \caption{Cars}
    \label{fig:3-c}
  \end{subfigure}
    \begin{subfigure}{0.32\linewidth}
    \includegraphics[width=\linewidth]{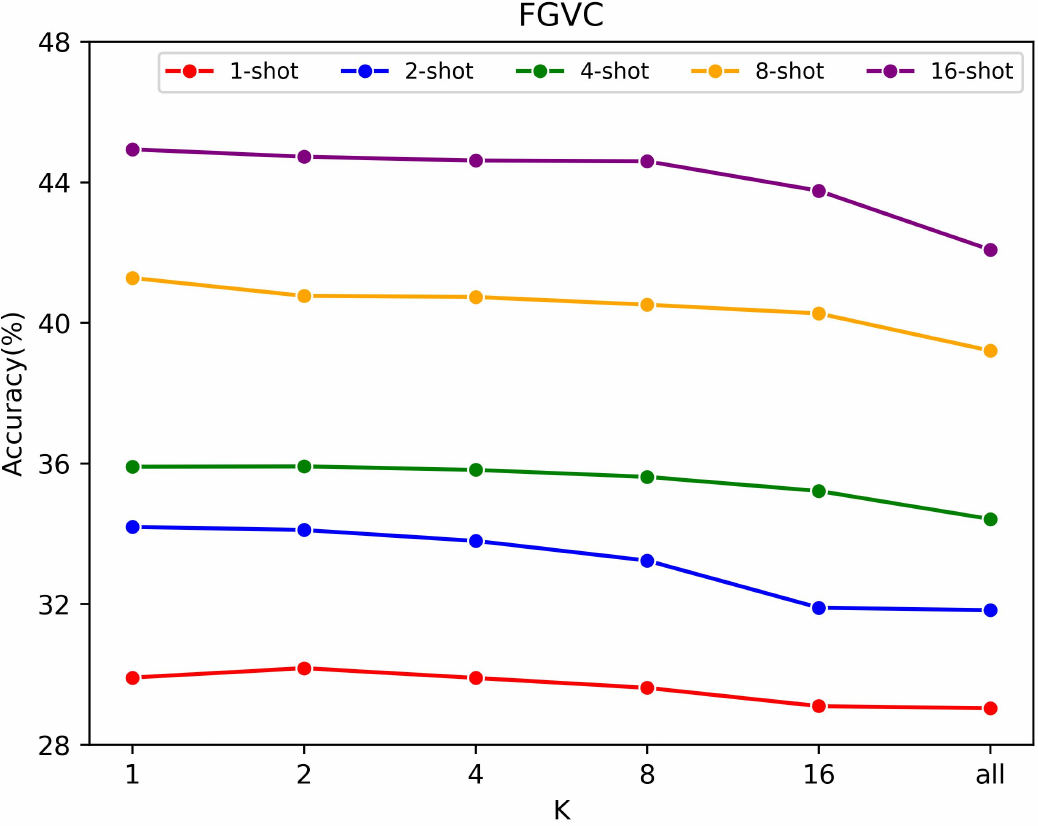}
    \caption{FGVC}
    \label{fig:3-f}
  \end{subfigure}
  \hfill
  \caption{Ablation study on $K$ values based on MaPLe \cite{khattak2023maple}. Note that we incorporate all unlabeled samples in test set for fair comparison.}
  \label{fig:3}
\end{figure}
Since KCL iteratively select the top $K$ mutual nearest neighbors from unlabeled samples for knowledge completion, the choice of $K$ value of becomes a key issue. Here we try different $K$ values ranging from 1 to full size of test set and present the corresponding classification results in Fig. \ref{fig:3}. We can find that when the total number of unlabeled samples remains unchanged, the smaller $K$, the better the performance. In general, when $K$ is small, the performance decrease is not significant; when $K$ becomes large, the performance decreases quickly. This is because smaller $K$ value leads to a smoother and robust knowledge completion process while larger $K$ leads to a rugged one. Therefore, in this paper, we report all experimental results based on $K$=1. 

\textbf{Number of unlabeled samples required for completion.}
\begin{figure}[t]
  \centering
  \begin{subfigure}{0.32\linewidth}
    \includegraphics[width=0.99\linewidth]{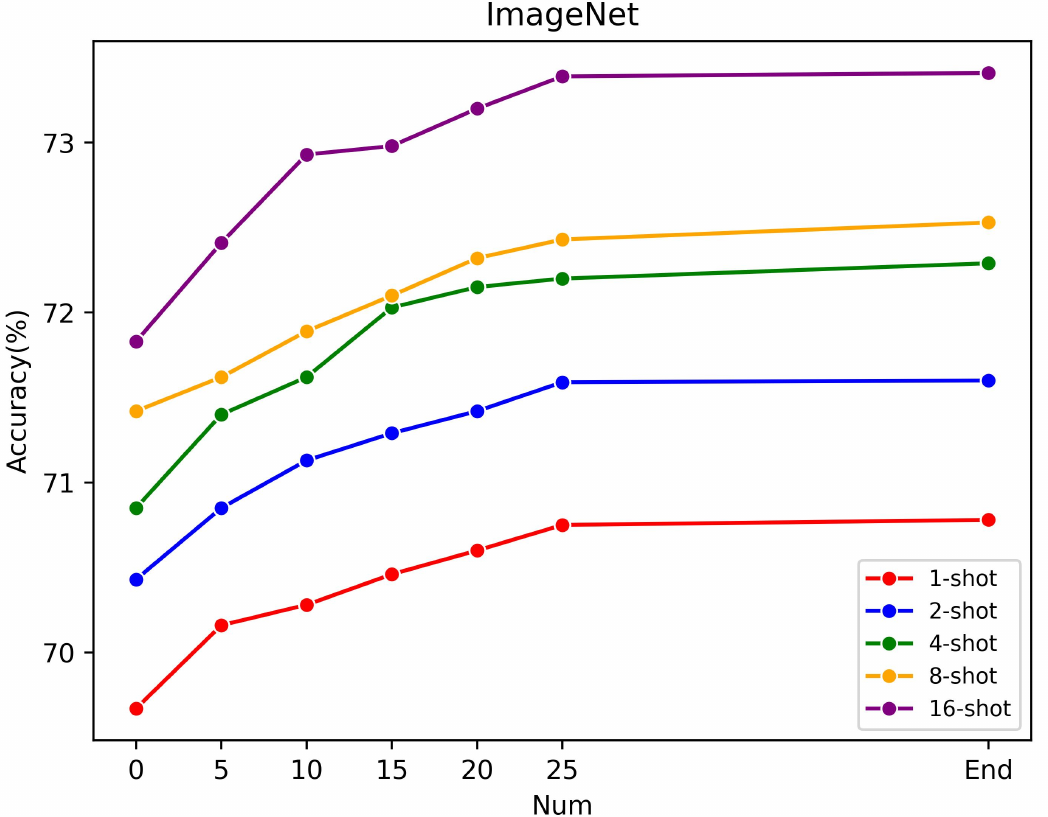}
    \caption{ImageNet}
    \label{fig:4-a}
  \end{subfigure}
  \begin{subfigure}{0.32\linewidth}
    \includegraphics[width=0.99\linewidth]{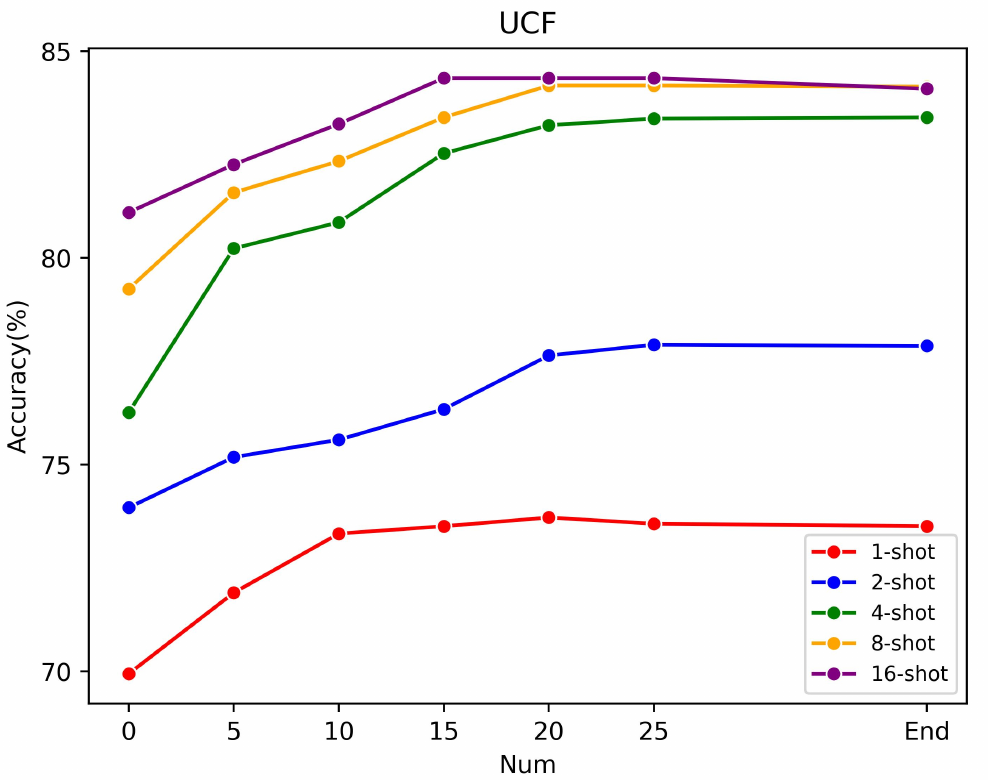}
    \caption{UCF}
    \label{fig:4-b}
  \end{subfigure}
   \begin{subfigure}{0.32\linewidth}
    \includegraphics[width=0.99\linewidth]{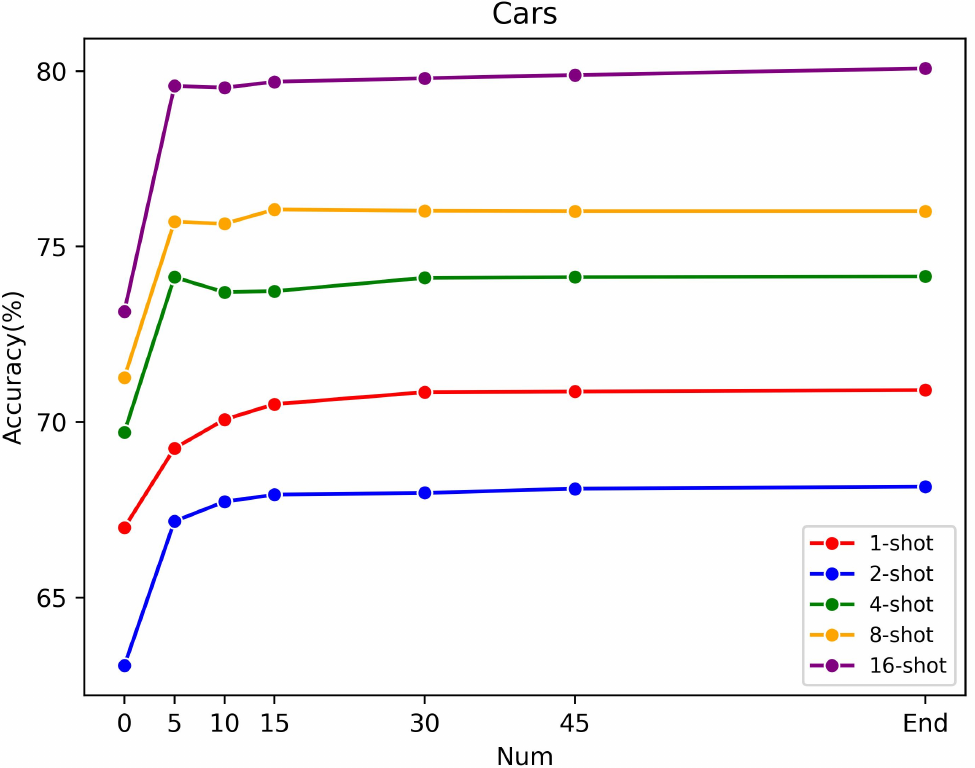}
    \caption{Cars}
    \label{fig:4-c}
  \end{subfigure}
  \hfill
  \caption{Ablation study on number of unlabeled samples required for knowledge completion. Note that we conduct experiments based on MaPLe \cite{khattak2023maple} under few-shot settings.}
  \label{fig:4}
\end{figure}
In this work, we propose KCL to take advantages of unlabeled samples for knowledge completion. However, in real applications, unlabeled samples can be infinite thus it is time-consuming to make use of all unlabeled samples for knowledge completion. Therefore, in this section, we take a close look at the number of unlabeled samples required for convergence. Specifically, we conduct experiments on three datasets to observe the relationships between classification performance and number of unlabeled samples (note that we choose $K$=1 for all datasets). From Fig. \ref{fig:4} we can see that as the number of unlabeled samples increases, classification performance increases rapidly at beginning and then levels off till the end. Moreover, we can notice that on relatively simpler fine-grained datasets like Cars, only 5-10 unlabeled samples per category are enough for visual knowledge completion; while on relatively harder datasets like ImageNet and UCF, nearly 20-25 samples per category are needed for complete visual knowledge. This phenomenon indicates that KCL has no need to take all unlabeled data into account, instead, only 25 samples per category are enough for achieving remarkable improvements, ensuring a high efficiency of KCL method. 

\textbf{Effectiveness of the designed confidence criterion.}
\begin{figure}[t]
  \centering
  \begin{subfigure}{0.32\linewidth}
    \includegraphics[width=\linewidth]{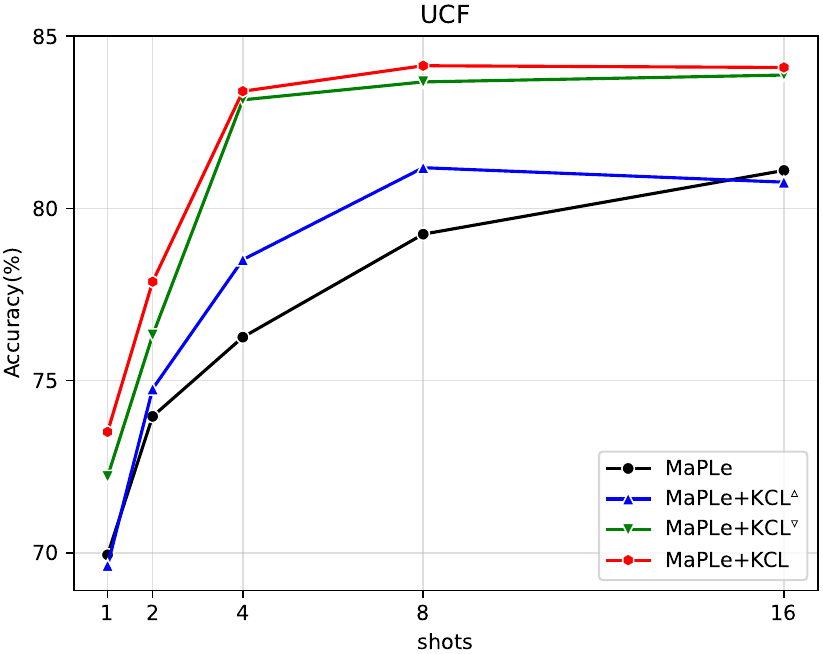}
    \caption{UCF}
    \label{fig:5-a}
  \end{subfigure}
  \begin{subfigure}{0.32\linewidth}
    \includegraphics[width=\linewidth]{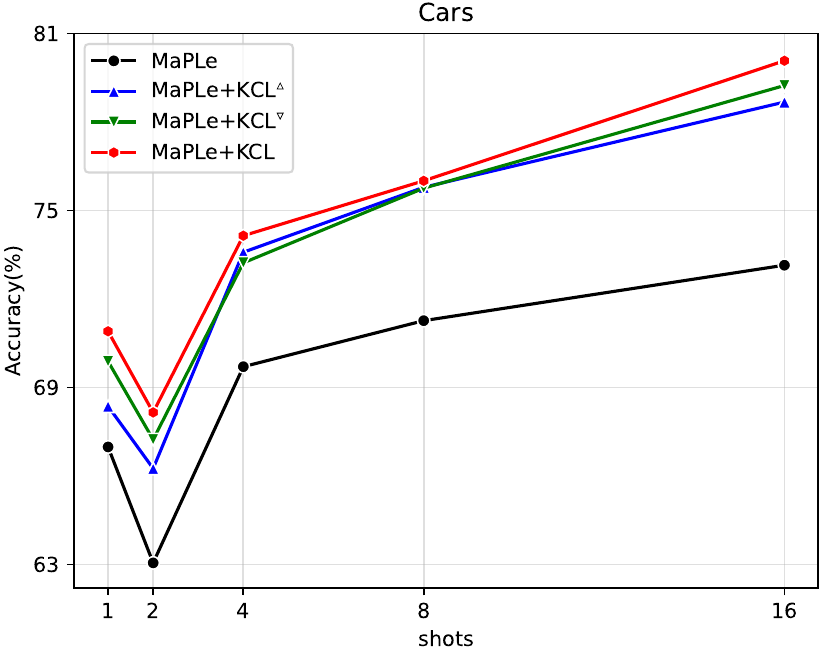}
    \caption{Cars}
    \label{fig:5-b}
  \end{subfigure}
   \begin{subfigure}{0.32\linewidth}
    \includegraphics[width=\linewidth]{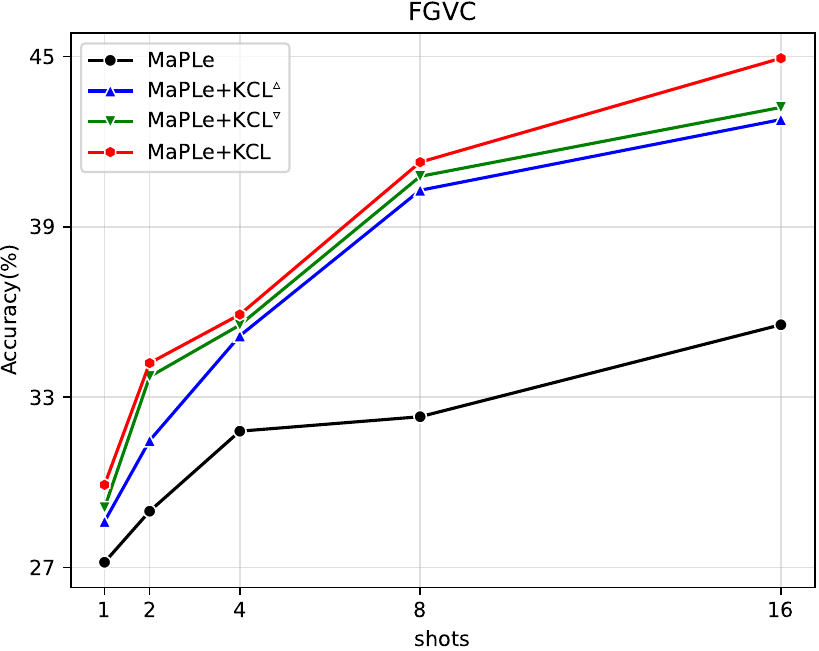}
    \caption{FGVC}
    \label{fig:5-c}
  \end{subfigure}
  \hfill
  \caption{Ablation study on the designed confidence criterion based on MaPLe \cite{khattak2023maple}. }
  \label{fig:5}
\end{figure}
In this paper, we select and incorporate unlabeled samples according to a designed confidence criterion based on mutual nearest neighbor theory \cite{gowda1979condensed}, which is a core content of KCL. Besides the remarkable improvements made by KCL, we would like to fully reveal the contribution of the designed confidence criterion. Specifically, we compare KCL with two types of degraded nearest neighbor (NN) strategies: (1) image-to-category NN strategy chooses the most similar samples to each category for knowledge completion, which is given by modifying Eq. \ref{eq5} to $\overline{\mathcal{M}}_{K}(S_i) = \{f_j \in \bm{f}|f_j \in \mathcal{N}_{K}^{\bm f}(S_i)\}$, termed as KCL$^{\triangledown}$; (2) category-to-image NN strategy makes choices following modified Eq. \ref{eq5}: $\widehat{\mathcal{M}}_{K}(S_i) = \{f_j \in \bm{f}| S_i \in \mathcal{N}_{K}^{\bm S}(f_j)\}$, termed as KCL$^{\vartriangle}$. As is shown in Fig. \ref{fig:5}, we conduct ablation study on three datasets based on MaPLe \cite{khattak2023maple} and we can observe that KCL always achieves the best performance among different criteria. This demonstrates the superiority of the MNNs-based confidence criterion, which can give the confidence selection process a double guarantee.

\textbf{Modalities for computing similarity matrix.}
\begin{figure}[t]
  \centering
  \begin{subfigure}{0.32\linewidth}
    \includegraphics[width=\linewidth]{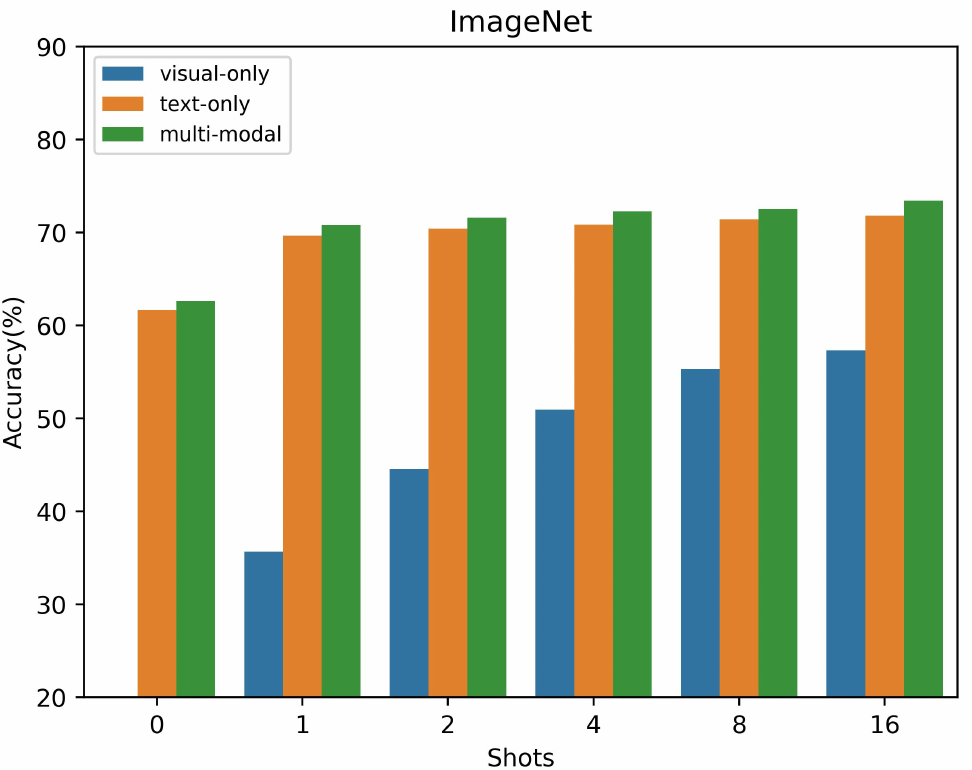}
    \caption{ImageNet}
    \label{fig:6-a}
  \end{subfigure}
  \begin{subfigure}{0.32\linewidth}
    \includegraphics[width=\linewidth]{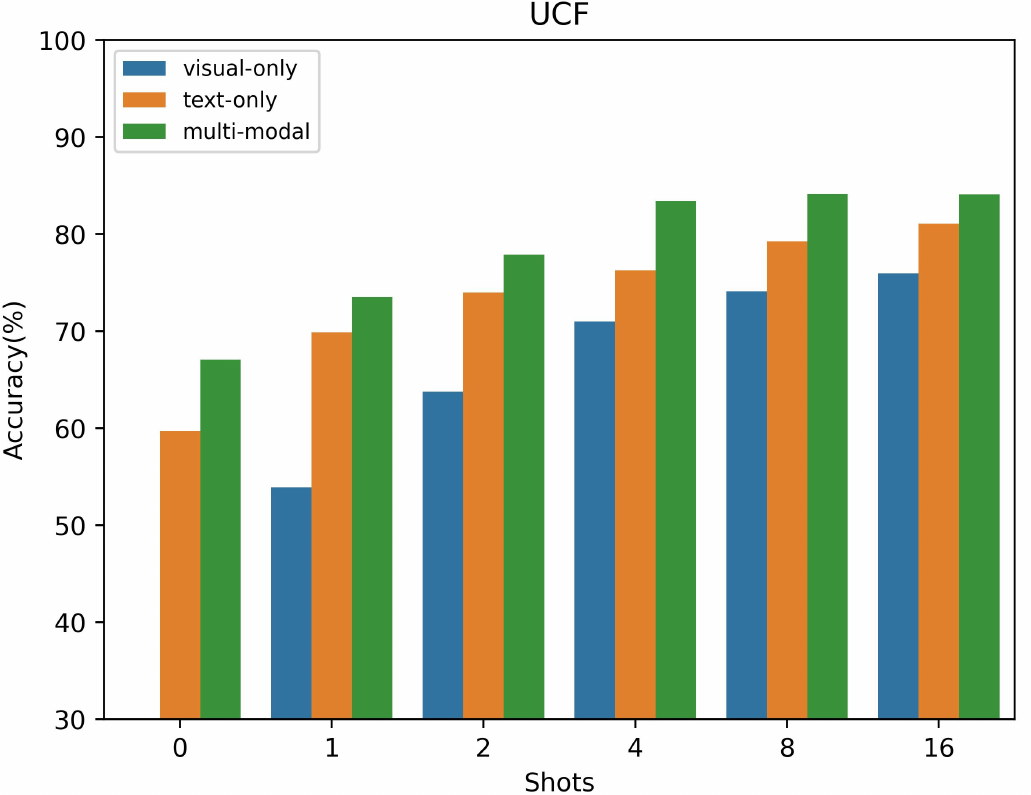}
    \caption{UCF}
    \label{fig:6-b}
  \end{subfigure}
   \begin{subfigure}{0.32\linewidth}
    \includegraphics[width=\linewidth]{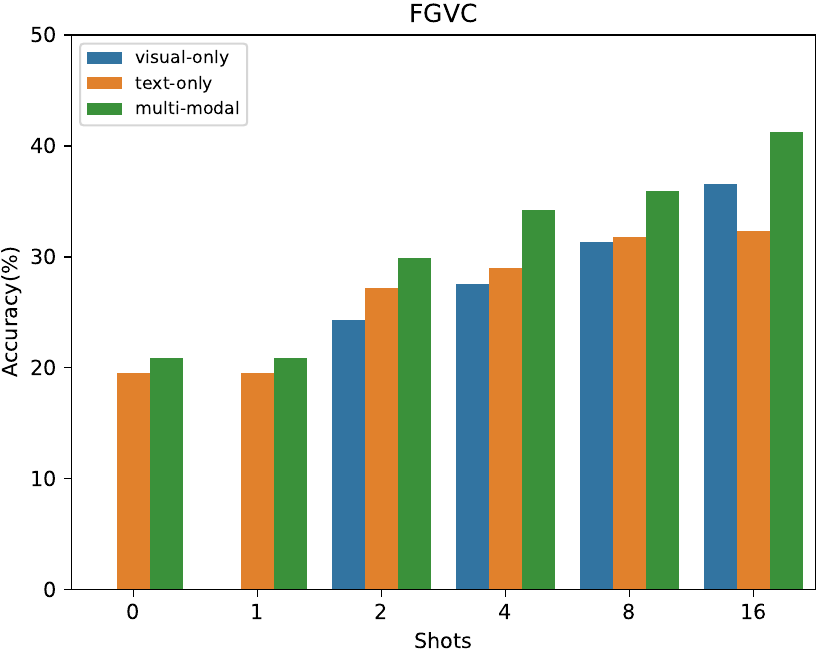}
    \caption{FGVC}
    \label{fig:6-c}
  \end{subfigure}
  \hfill
  \caption{Ablation study on modalities chosen for computing similarity matrix. Note that we report results based on MaPLe \cite{khattak2023maple} under both zero-shot and few-shot settings.}
  \label{fig:6}
\end{figure}
As is shown in Eq. \ref{eq6} and Eq. \ref{eq7}, we combine a visual similarity matrix and a textual similarity matrix as a multi-modal similarity matrix. To prove the effectiveness of visual and textual knowledge for predicting high-confidence unlabeled samples, we conduct comparative experiments on visual-only and text-only settings. As is shown in Fig. \ref{fig:6}, we can find that multi-modal similarity matrix achieves the best performance in all cases. Besides, performance under text-only setting is better than that under visual-only setting in most cases except for 8-shot and 16-shot FGVC tasks. This may be because FGVC is a professional fine-grained dataset of various rare aircrafts, which may exceed the scope of CLIP's generalization capabilities, thus requiring more assistance from visual information.

\section{Discussion}
\label{dis}

\begin{figure*}[t]
\begin{minipage}[t!]{0.41\linewidth}
    \centering
    \includegraphics[width=\linewidth]{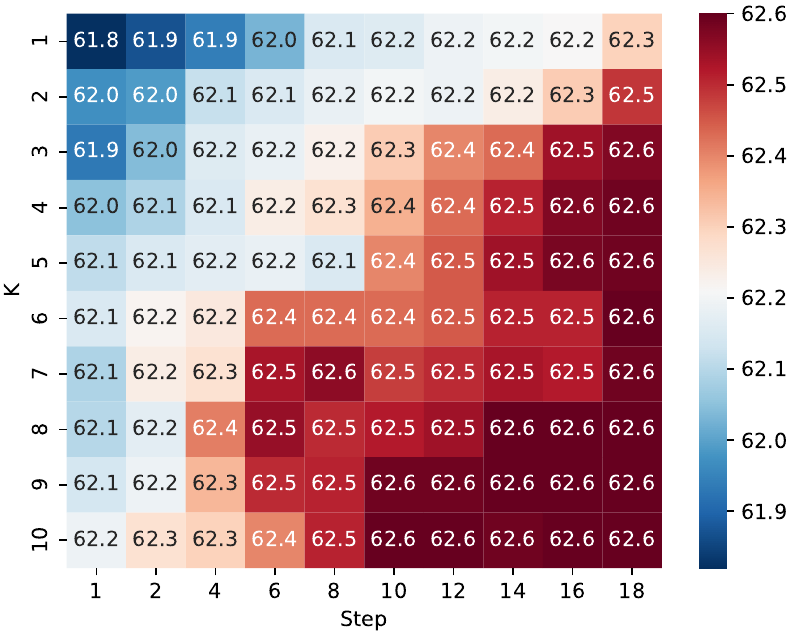}
    \figcaption{Classification performance (\%) with different $K$ and iteration steps on ImageNet \cite{deng2009imagenet}.}
    \label{fig:6}
\end{minipage}\qquad
\begin{minipage}[t!]{0.53\linewidth}
    \begin{adjustbox}{width=\linewidth}
        \begin{tabular}{lc|c}
        \toprule[1pt]
         \textbf{Method} & \textbf{Completion Time}   &\textbf{Accuracy (\%)}\\
        \midrule
        \textbf{CLIP}$^{*}$ \cite{radford2021learning} & --         & 60.32 \\
        \textbf{VisDesc}$^{*}$ \cite{menon2022visual}  & $\sim$3.0s   & 59.68 \\    
        \textbf{CuPL} $^{*}$\cite{pratt2023does}       & $\sim$3.0s   &  61.64 \\
        \textbf{SuS-X-SD}$^{*}$\cite{udandarao2023sus} & $\sim$60.0s  & 61.84 \\
        \textbf{SuS-X-LC}$^{*}$\cite{udandarao2023sus} & $\sim$2.0s   & 61.89 \\
        \midrule
        \textbf{KCL$_{K=1,Step=2}$}$\ddagger$ & $\sim$\textbf{0.8s}   & 61.87 \\
        \textbf{KCL$_{K=1,Step=4}$}$\ddagger$ & $\sim$\textbf{1.5s}   & \textbf{61.94} \\
        \textbf{KCL$_{K=3,Step=18}$}$\ddagger$ & $\sim$41.4s  & \textbf{62.60} \\
        
    \bottomrule[1pt]
        \end{tabular}
    	\end{adjustbox}
	\tabcaption{Classification performance (\%) and completion time cost for one ImageNet \cite{deng2009imagenet} class. $\ddagger$ denotes our implementation on single V100-32G GPU and $^{*}$ denotes results reported in \cite{udandarao2023sus} with single A100-80G GPU.}
	\label{tab:3}
\end{minipage}
\end{figure*}
\textbf{Q: KCL requires selecting and incorporating high-confidence samples iteratively, can it stay efficient compared to other methods? A: Yes.}

In this work, we propose to complement visual knowledge by repeatedly selecting and incorporating unlabeled samples with high confidence, which may brings complex operations and high computational costs. We conduct in-depth experiments on zero-shot ImageNet \cite{deng2009imagenet} to prove the efficiency of KCL. As is shown in Fig. \ref{fig:6}, we first present classification accuracy under different $K$ and iteration steps. We can see that when the number of unlabeled samples ($K \times \text{step}$) get larger, visual knowledge completion can reach convergence faster. Moreover, we also compare the time costs of KCL with other knowledge completion methods on zero-shot tasks, in which VisDesc \cite{menon2022visual} and CuPL \cite{pratt2023does} make use of external knowledge bases (e.g. GPT-3) to complement textual knowledge; while SuS-X \cite{udandarao2023sus} utilizes generative models and auxiliary databases to complement visual knowledge. From Table \ref{tab:3} we can observe that KCL can achieve higher classification accuracy with less completion time cost, achieving a good balance between performance and efficiency.

\noindent\textbf{Q: KCL takes advantage of unlabeled data, does it duplicate with semi-supervised learning or transductive inference? A: No.}

Semi-supervised learning \cite{zhou2021semi, ren2018meta} is one of the important approaches for few-shot and zero-shot learning, which takes advantages of external unlabeled databases by pseudo labeling. Differently, KCL does not require access to auxiliary databases, which leverages the power of pre-trained CLIP and directly makes use of test data with high confidence. Besides, the core concept of transductive inference \cite{vapnik2013nature,liu2019learning,li2022transductive} is to predict test data as a whole without learning a rule first like common inductive inference. Differently, KCL goes beyond transduction and inherits an advanced concept of \textbf{\emph{selective inference}} \cite{vapnik200624}, which selects among the unlabeled test samples the most confident ones. KCL takes it further by utilizing the selected samples to conduct visual knowledge completion for alleviating the bias caused by data scarcity. Therefore, KCL is better prepared for difficult test samples with low confidence than conventional transductive inference, especially in low-shot tasks. As is shown in Table \ref{tab:4}, KCL outperforms transductive label propagation (TLP) \cite{liu2019learning,li2022transductive}, which is a common practice for transductive inference, demonstrating the above points. 

\begin{table}[t]
\centering
\caption{Classification results (\%) of KCL and transductive label propagation (TLP).}
\label{tab:4}
\resizebox{\textwidth}{!}{
\begin{tabular}{lccccc|ccccc}
\toprule[1pt]
\multirow{2}{*}{\textbf{Method}} & \multicolumn{5}{c|}{\textbf{UCF} }                                                        & \multicolumn{5}{c}{\textbf{FGVC} }                                                        \\ \cmidrule{2-11} 
                                 & {1-shot} & {2-shot} & {4-shot} & {8-shot} & {16-shot} & {1-shot} & {2-shot} & {4-shot} & {8-shot} & {16-shot} \\ \midrule
\textbf{MaPLe+TLP}            & 71.03           & 75.10           & 78.43           & 81.34           & 82.98            & 28.14           & 31.05           & 34.05           & 35.19           & 39.57            \\
\textbf{MaPLe+KCL}               & \textbf{73.51}  & \textbf{77.87}  & \textbf{83.40}   & \textbf{84.14}  & \textbf{84.09}   & \textbf{29.91}  & \textbf{34.20}   & \textbf{35.91}  & \textbf{41.28}  & \textbf{44.94}   \\ \bottomrule[1pt]
\end{tabular}}
\end{table}

\noindent\textbf{Q: Any limitations or subsequent improvements for KCL? A: Yes.}

Despite the remarkable performance achieved by KCL, there are still some limitations need to be improved in the future: (1) The knowledge completion process of KCL highly dependents on pre-trained CLIP models, which assumes CLIP has been exposed to everything during pre-training stage. When facing something extremely rare, our method may fail to transfer effectively, which is also a common problem in existing low-shot learning methods \cite{zhou2022learning, khattak2023maple, udandarao2023sus}. (2) Following most existing approaches for CLIP-based transfer learning \cite{guo2023calip, zhou2022conditional, zhu2023not}, KCL still targets at solving closed-set low-shot classification problems, which is not consistent with real application scenarios where open-set and incremental classes exist \cite{miyai2024locoop}. KCL may take a high risk of failure under such real settings, being unable to identify out-of-distribution samples or catastrophically forget previously learned concepts.

\section{Conclusion}
In this work, we figure out the fundamental factor that constrains few-shot learning performance of CLIP: the biased visual knowledge caused by narrow distribution of few shots. To tackle this problem, we propose an iterative visual knowledge completion method (KCL) to make use of unlabeled test data properly with a designed confidence criterion and further extend it into zero-shot learning tasks. Unlike existing methods, KCL conducts visual knowledge completion without access to auxiliary or synthetic data. Extensive experiments on 11 benchmark datasets demonstrate the effectiveness of KCL in both zero-shot and few-shot settings, acting as a plug-and-play module to existing state-of-the-arts. Ablation study and further analyses also prove the efficiency and rationality of KCL method. Our future work will focus on adapting KCL to real applications with open-set and continuous categories.

%
%

\newpage
\appendix
\paragraph{{\rm {\bf Supplementary Materials}}}

\section{Dataset Details}
We provide the information of validation and testing splits for 11 benchmark datasets in Table \ref{tab:datasets}. The introductions of all datasets are presented as follows:

\begin{table}[h]

    \caption{Brief introduction to 11 benchmark datasets.}
    \label{tab:datasets}
    \resizebox{\textwidth}{!}{
    \begin{tabular}{l l l l c}
    \toprule[1pt]
    \textbf{Dataset} & \textbf{Classes} & \textbf{Val} & \textbf{Test} & \textbf{Introduction} \\
    \midrule
    \textbf{ImageNet} & 1,000 &  50,000 & 50,000 & Images organized according to WordNet hierarchy \\
    \textbf{Caltech} & 100  & 1,649 & 2,465 & Dataset contains pictures of objects \\ 
    \textbf{Pets} & 37  & 736 & 3,669 &  Dataset consists of different pets \\
    \textbf{Cars} & 196  & 1,635 & 8,041 &  Cars images taken from the rear\\
    \textbf{Flowers} & 102  & 1,633 & 2,463 & Images of flowers commonly occurring in UK \\
    \textbf{Food} & 101  & 20,200 & 30,300 &  Images of different food organized by types\\
    \textbf{FGVC} & 100  & 3,333 & 3,333 &  Benchmark dataset for categorization of aircraft.\\
    \textbf{SUN} & 397  & 3,970 & 19,850 &  Benchmark dataset in scene understanding \\
    \textbf{DTD} & 47  & 1,128 & 1,692 & Collection of textural images in the wild \\
    \textbf{EuroSAT} & 10  & 5,400 & 8,100 &  Sentinel-2 satellite images covering 13 spectral bands\\
    \textbf{UCF} & 101  & 1,898 & 3,783 &  Video clips collected from Youtube \\
    
    \bottomrule[1pt]
    \end{tabular}
    }
\end{table}
\begin{enumerate}
    \item \textbf{ImageNet}: In this work, we use one of the most highly used subset of ImageNet namely the "ImageNet Large Scale Visual Recognition Challenge (ILSVRC) 2012-2017 image classification and localization dataset" which is also referred to in the research literature as ImageNet-1K or ILSVRC2017, reflecting the original ILSVRC challenge that involved 1,000 classes.\footnote[1]{https://en.wikipedia.org/wiki/ImageNet}
    \item \textbf{UCF}: UCF is an action recognition data set of realistic action videos, collected from YouTube with 101 action categories. The videos in 101 action categories are grouped into 25 groups, where each group can consist of 4-7 videos of an action. \footnote[2]{https://www.crcv.ucf.edu/data/UCF101.php} Following CoOp\cite{zhou2022learning}, we consider the middle frame of each video as our image sample.
    \item \textbf{DTD}: DTD is a texture database, consisting of 5640 images, organized according to a list of 47 categories with 120 images per category. The images were collected from Google and Flickr by entering the proposed attributes and related terms as search queries. The images were annotated using Amazon Mechanical Turk in several iterations. \footnote[3]{https://www.robots.ox.ac.uk/\textasciitilde vgg/data/dtd/}
    \item \textbf{FGVC}: FGVC contains 100 images for each of different aircraft model variants, most of which are airplanes. The (main) aircraft in each image is annotated with a tight bounding box and a hierarchical airplane model label. Aircraft models are organized in a four-levels hierarchy: model, variant, family and manufacturer. \footnote[4]{{https://www.robots.ox.ac.uk/\textasciitilde vgg/data/fgvc-aircraft/}}
    \item \textbf{Flowers}: Oxford Flower is an image classification dataset consisting of 102 flower categories. The flowers chosen to be flower commonly occurring in the United Kingdom. Each class consists of 40-258 images. \footnote[5]{https://paperswithcode.com/dataset/oxford-102-flower}
    \item \textbf{Cars}: The Stanford Cars dataset consists of 196 classes of cars with a total of 16,185 images. \footnote[6]{https://paperswithcode.com/dataset/stanford-cars}
    \item \textbf{Eurosat}\cite{helber2019eurosat}: Eurosat is a dataset and deep learning benchmark for land use and land cover classification. The dataset is based on Sentinel-2 satellite images covering 13 spectral bands, consisting of 10 classes with in total 27,000 labeled images.
    \item \textbf{SUN}: The database contains 108,753 images of 397 categories, used in the Scene UNderstanding (SUN) benchmark. The number of images varies across categories, but there are at least 100 images per category. \footnote[7]{https://paperswithcode.com/dataset/sun397}
    \item \textbf{Pets}\cite{krause20133d}: The Oxford-Pet Dataset has 37 categories with roughly 200 images in each class and each image has an associated ground truth annotation of breed, head ROI, and pixel level segmentation.
    \item \textbf{Caltech}: Caltech consists of pictures of objects belonging to 101 classes as well as a background clutter class. Each image is labelled with a single object. Each class contains roughly 40-800 samples with totally 9,000 images. \footnote[8]{https://www.tensorflow.org/datasets/catalog/caltech101}Note that we discard the "BACKGROUND Google" and "Faces easy classes" from Caltech101 dataset as described in \cite{zhou2022learning}. 
    \item \textbf{Food}: This dataset consists of 101 food categories, with totally 101,000 images. All images were rescaled to have a maximum side length of 512 pixels.  \footnote[9]{https://www.tensorflow.org/datasets/catalog/food101}

\end{enumerate}

\section{Overall Experimental Results of Few-shot Learning}

As is shown in Table \ref{tab:overall}, we present the overall improvements achieved by KCL on five representative methods under 1, 2, 4, 8 and 16-shot settings, including APE \cite{zhu2023not}. We can find that, as a training-free plug-and-play module, KCL can significantly boost existing methods from different types in almost all cases, which demonstrates its effectiveness in few-shot learning scenarios.

\begin{table}[htbp]
\centering
\caption{Overall experimental results of few-shot learning on 11 benchmark datasets.}
\label{tab:overall}
\resizebox{0.97\textwidth}{!}{
\begin{tabular}{l|c|ccccccccccc|c}
\toprule[1pt]
 {\textbf{Methods}} & \rotatebox{60}{\textbf{Shots}} & \rotatebox{60}{\textbf{ImageNet}} & \rotatebox{60}{\textbf{UCF}} & \rotatebox{60}{\textbf{DTD}} & \rotatebox{60}{\textbf{FGVC}} & \rotatebox{60}{\textbf{Flowers}} & \rotatebox{60}{\textbf{Cars}} & \rotatebox{60}{\textbf{Eurosat}} & \rotatebox{60}{\textbf{SUN}} & \rotatebox{60}{\textbf{Pets}} & \rotatebox{60}{\textbf{Caltech}} & \rotatebox{60}{\textbf{Food}} & \rotatebox{60}{\textbf{Average}} \\ \midrule

\textbf{CoOp} \cite{zhou2022learning} &\multirow{10}{*}{\textbf{1}}       & 55.70                              & 64.92                        & 53.25                        & 21.18                         & 81.81                            & 53.89                         & 62.56                            & 60.68                           & 81.60                          & 90.06                            & 73.42                         & 63.55                            \\
$\bm\Delta \textbf{+KCL}$(Ours) &       & {\color[HTML]{009901} +2.84}      & {\color[HTML]{009901} +4.73} & {\color[HTML]{009901} +4.73} & {\color[HTML]{009901} +0.18}  & {\color[HTML]{009901} +4.22}     & {\color[HTML]{009901} +3.40}   & {\color[HTML]{009901} +6.75}     & {\color[HTML]{009901} +2.77}    & {\color[HTML]{009901} +3.11}  & {\color[HTML]{009901} +0.49}     & {\color[HTML]{009901} +2.76}  & {\color[HTML]{009901} +3.27}     \\ \cmidrule{1-1}\cmidrule{3-14}
\textbf{CLIP-Adapter} \cite{gao2023clip} &      & 61.94                             & 64.58                        & 51.60                         & 21.27                         & 72.03                            & 58.77                         & 61.23                            & 63.83                           & 85.61                         & 90.10                             & 76.80                          & 64.34                            \\
$\bm\Delta \textbf{+KCL}$(Ours) &       & {\color[HTML]{009901} +1.11}      & {\color[HTML]{009901} +6.90}  & {\color[HTML]{009901} +5.97} & {\color[HTML]{009901} +1.47}  & {\color[HTML]{009901} +10.63}    & {\color[HTML]{009901} +1.67}  & {\color[HTML]{009901} +0.99}     & {\color[HTML]{009901} +1.83}    & {\color[HTML]{009901} +1.17}  & {\color[HTML]{009901} +1.06}     & {\color[HTML]{009901} +0.69}  & {\color[HTML]{009901} +3.04}     \\ \cmidrule{1-1}\cmidrule{3-14}
\textbf{Tip-Adapter} \cite{zhang2022tip} &      & 61.83                             & 63.86                        & 51.12                        & 20.94                         & 72.63                            & 58.60                          & 57.68                            & 54.08                           & 85.25                         & 89.25                            & 77.57                         & 63.89                            \\
$\bm\Delta \textbf{+KCL}$(Ours) &       & {\color[HTML]{009901} +0.71}      & {\color[HTML]{009901} +6.24} & {\color[HTML]{009901} +5.44} & {\color[HTML]{009901} +0.63}  & {\color[HTML]{009901} +9.79}     & {\color[HTML]{009901} +2.74}  & {\color[HTML]{009901} +7.05}     & {\color[HTML]{009901} +1.50}     & {\color[HTML]{009901} +1.20}   & {\color[HTML]{009901} +1.34}     & {\color[HTML]{009901} +0.61}  & {\color[HTML]{009901} +3.38}     \\ \cmidrule{1-1}\cmidrule{3-14}
\textbf{APE} \cite{zhu2023not} &      & 62.00                             & 63.86                        & 52.90                        & 20.97                         & 79.33                            & 59.89                          &  61.15                            & 64.18                           & 84.88                         & 90.22                            & 77.27                         &  65.15                            \\
$\bm\Delta \textbf{+KCL}$(Ours) &       & {\color[HTML]{009901} +1.20}      & {\color[HTML]{009901} +6.03} & {\color[HTML]{009901} +4.37} & {\color[HTML]{009901} +2.10}  & {\color[HTML]{009901} +3.01}     & {\color[HTML]{009901} +2.61}  & {\color[HTML]{009901} +10.71}     & {\color[HTML]{009901} +1.44}     & {\color[HTML]{009901} +1.68}   & {\color[HTML]{009901} +0.21}     & {\color[HTML]{009901} +0.91}  & {\color[HTML]{009901} +3.11}     \\ \cmidrule{1-1}\cmidrule{3-14}

\textbf{MaPLe} \cite{khattak2023maple} &      & 69.67                             & 69.94                        & 41.31                        & 27.18                         & 77.30                             & 66.99                         & 60.75                            & 70.05                           & 91.55                         & 93.06                            & 84.12                         & 68.35                            \\
$\bm\Delta \textbf{+KCL}$(Ours) &       & {\color[HTML]{009901} +1.11}      & {\color[HTML]{009901} +3.57} & {\color[HTML]{009901} +6.56} & {\color[HTML]{009901} +2.73}  & {\color[HTML]{009901} +12.87}    & {\color[HTML]{009901} +3.92}  & {\color[HTML]{009901} +6.78}     & {\color[HTML]{009901} +1.29}    & {\color[HTML]{009901} +0.27}  & {\color[HTML]{009901} +0.37}     & {\color[HTML]{009901} +0.22}  & {\color[HTML]{009901} +3.60}     \\ \midrule

\textbf{CoOp} \cite{zhou2022learning}&\multirow{10}{*}{\textbf{2}}& 57.36                             & 68.44                        & 55.85                        & 25.11                         & 84.94                            & 60.17                         & 65.49                            & 62.03                           & 84.19                         & 89.49                            & 73.58                         & 66.05                            \\
$\bm\Delta \textbf{+KCL}$(Ours) &       & {\color[HTML]{009901} +1.15}      & {\color[HTML]{009901} +4.97} & {\color[HTML]{009901} +0.53} & {\color[HTML]{CB0000} -0.18}  & {\color[HTML]{009901} +3.61}     & {\color[HTML]{009901} +1.03}  & {\color[HTML]{009901} +7.52}     & {\color[HTML]{009901} +3.17}    & {\color[HTML]{009901} +1.36}  & {\color[HTML]{009901} +0.17}     & {\color[HTML]{009901} +2.28}  & {\color[HTML]{009901} +2.32}     \\ \cmidrule{1-1}\cmidrule{3-14}
\textbf{CLIP-Adapter} \cite{gao2023clip}&      & 61.84                             & 66.88                        & 56.97                        & 21.87                         & 73.12                            & 58.92                         & 62.23                            & 64.11                           & 87.33                         & 90.83                            & 77.34                         & 65.58                            \\
$\bm\Delta \textbf{+KCL}$(Ours) &       & {\color[HTML]{009901} +1.00}      & {\color[HTML]{009901} +2.93} & {\color[HTML]{009901} +3.49} & {\color[HTML]{009901} +0.84}  & {\color[HTML]{009901} +3.25}     & {\color[HTML]{009901} +1.12}  & {\color[HTML]{009901} +5.89}     & {\color[HTML]{009901} +1.21}    & {\color[HTML]{009901} +0.62}  & {\color[HTML]{009901} +0.08}     & {\color[HTML]{009901} +0.54}  & {\color[HTML]{009901} +1.90}     \\ \cmidrule{1-1}\cmidrule{3-14}
\textbf{Tip-Adapter} \cite{zhang2022tip} &      & 62.01                             & 66.56                        & 54.85                        & 23.10                         & 76.49                            & 60.87                         & 59.07                            & 65.49                           & 84.25                         & 89.49                            & 77.35                          & 65.32                            \\
$\bm\Delta \textbf{+KCL}$(Ours) &       & {\color[HTML]{009901} +0.84}      & {\color[HTML]{009901} +4.57} & {\color[HTML]{009901} +2.36} & {\color[HTML]{009901} +0.48}  & {\color[HTML]{009901} +8.12}     & {\color[HTML]{009901} +1.40}   & {\color[HTML]{009901} +9.10}      & {\color[HTML]{009901} +1.16}    & {\color[HTML]{009901} +2.56}  & {\color[HTML]{009901} +0.53}     & {\color[HTML]{009901} +0.45}  & {\color[HTML]{009901} +2.87}     \\ \cmidrule{1-1}\cmidrule{3-14}

\textbf{APE} \cite{zhu2023not} &      & 62.40                             & 66.22                       & 57.04                        & 22.71                         & 83.64                           & 60.91                          &  63.37                           & 65.88                           & 83.95                         & 90.02                           & 77.61                         &  66.79                            \\
$\bm\Delta \textbf{+KCL}$(Ours) &       & {\color[HTML]{009901} +1.00}      & {\color[HTML]{009901} +4.62} & {\color[HTML]{009901} +3.89} & {\color[HTML]{CB0000} -0.51}  & {\color[HTML]{009901} +4.75}     & {\color[HTML]{009901} +1.89}  & {\color[HTML]{009901} +4.78}     & {\color[HTML]{009901} +0.77}     & {\color[HTML]{009901} +2.42}   & {\color[HTML]{009901} +1.18}     & {\color[HTML]{009901} +0.64}  & {\color[HTML]{009901} +2.31}     \\ \cmidrule{1-1}\cmidrule{3-14}

\textbf{MaPLe} \cite{khattak2023maple} &      & 70.43                             & 73.96                        & 53.66                        & 28.98                         & 79.54                            & 63.06                         & 57.89                            & 70.35                           & 91.09                         & 93.87                            & 86.33                         & 69.92                            \\
$\bm\Delta \textbf{+KCL}$(Ours) &       & {\color[HTML]{009901} +1.17}      & {\color[HTML]{009901} +3.91} & {\color[HTML]{009901} +4.79} & {\color[HTML]{009901} +5.22}  & {\color[HTML]{009901} +12.75}    & {\color[HTML]{009901} +5.10}   & {\color[HTML]{009901} +20.64}    & {\color[HTML]{009901} +2.43}    & {\color[HTML]{009901} +0.46}  & {\color[HTML]{009901} +0.25}     & {\color[HTML]{009901} +0.58}  & {\color[HTML]{009901} +5.20}     \\ \midrule

\textbf{CoOp} \cite{zhou2022learning} & \multirow{10}{*}{\textbf{4}} & 58.86                             & 70.71                        & 61.64                        & 26.37                         & 89.77                            & 64.71                         & 72.70                             & 65.82                           & 87.24                         & 91.48                            & 74.91                         & 69.47                            \\
$\bm\Delta \textbf{+KCL}$(Ours) &       & {\color[HTML]{009901} +1.64}      & {\color[HTML]{009901} +4.28} & {\color[HTML]{009901} +0.18} & 0.00                             & {\color[HTML]{009901} +1.50}      & {\color[HTML]{009901} +0.19}  & {\color[HTML]{009901} +4.29}     & {\color[HTML]{009901} +0.95}    & {\color[HTML]{009901} +0.39}  & {\color[HTML]{CB0000} -0.04}     & {\color[HTML]{009901} +1.20}   & {\color[HTML]{009901} +1.32}     \\ \cmidrule{1-1}\cmidrule{3-14}
\textbf{CLIP-Adapter} \cite{gao2023clip}&      & 62.24                             & 69.15                        & 57.98                        & 22.20                          & 82.01                            & 60.81                         & 73.27                            & 66.74                           & 87.52                         & 91.16                            & 77.82                         & 68.26                            \\
$\bm\Delta \textbf{+KCL}$(Ours) &       & {\color[HTML]{009901} +0.98}      & {\color[HTML]{009901} +2.94} & {\color[HTML]{CB0000} -0.12} & {\color[HTML]{009901} +1.71}  & {\color[HTML]{009901} +6.42}     & {\color[HTML]{009901} +2.57}  & {\color[HTML]{009901} +5.15}     & {\color[HTML]{009901} +1.16}    & {\color[HTML]{009901} +0.35}  & {\color[HTML]{009901} +0.73}     & {\color[HTML]{009901} +0.41}  & {\color[HTML]{009901} +2.02}     \\ \cmidrule{1-1}\cmidrule{3-14}
\textbf{Tip-Adapter} \cite{zhang2022tip} &      & 62.16                             & 68.46                        & 59.22                        & 23.34                         & 82.34                             & 62.82                         & 69.27                            & 65.57                            & 84.79                          & 90.34                            & 77.25                         & 67.77                            \\
$\bm\Delta \textbf{+KCL}$(Ours) &       & {\color[HTML]{009901} +0.49}      & {\color[HTML]{009901} +3.18} & {\color[HTML]{009901} +1.60}  & {\color[HTML]{CB0000} -0.09}  & {\color[HTML]{009901} +4.26}     & {\color[HTML]{009901} +1.02}  & {\color[HTML]{009901} +2.74}     & {\color[HTML]{009901} +1.13}    & {\color[HTML]{009901} +1.61}  & {\color[HTML]{009901} +0.49}     & {\color[HTML]{009901} +0.49}  & {\color[HTML]{009901} +1.53}     \\ \cmidrule{1-1}\cmidrule{3-14}

\textbf{APE} \cite{zhu2023not} &      & 62.51                            & 68.76                       & 60.53                        & 24.61                         & 87.18                           & 64.11                          &  71.57                           & 66.99                           & 86.45                         & 91.63                           & 77.40                         &  69.33                            \\
$\bm\Delta \textbf{+KCL}$(Ours) &       & {\color[HTML]{009901} +1.10}      & {\color[HTML]{009901} +5.60} & {\color[HTML]{009901} +1.23} & {\color[HTML]{009901} +0.71}  & {\color[HTML]{009901} +2.26}     & {\color[HTML]{009901} +1.40}  & {\color[HTML]{009901} +4.31}     & {\color[HTML]{009901} +0.47}     & {\color[HTML]{CB0000} -0.40}   & {\color[HTML]{CB0000} -1.04}     & {\color[HTML]{009901} +0.84}  & {\color[HTML]{009901} +1.41}     \\ \cmidrule{1-1}\cmidrule{3-14}

\textbf{MaPLe} \cite{khattak2023maple} &      & 70.85                             & 76.26                        & 58.51                        & 31.80                          & 84.49                            & 69.71                         & 62.40                             & 72.00                              & 92.94                         & 94.44                            & 86.27                         & 72.69                            \\
$\bm\Delta \textbf{+KCL}$(Ours) &       & {\color[HTML]{009901} +1.44}      & {\color[HTML]{009901} +7.14} & {\color[HTML]{009901} +5.44} & {\color[HTML]{009901} +4.11}  & {\color[HTML]{009901} +9.26}     & {\color[HTML]{009901} +4.44}  & {\color[HTML]{009901} +16.59}    & {\color[HTML]{009901} +2.08}    & {\color[HTML]{009901} +0.03}  & {\color[HTML]{009901} +1.02}     & {\color[HTML]{009901} +0.14}  & {\color[HTML]{009901} +4.69}     \\
\midrule

\textbf{CoOp} \cite{zhou2022learning}&\multirow{10}{*}{\textbf{8}}& 60.30                             & 74.76                       & 64.78                        & 30.06                        & 93.34                            & 69.83                         & 78.60                            & 68.03                          & 86.67                         & 92.13                            & 75.92                         & 72.22                            \\
$\bm\Delta \textbf{+KCL}$(Ours) &       & {\color[HTML]{009901} +0.96}      & {\color[HTML]{009901} +3.85} & {\color[HTML]{009901} +0.65} & {\color[HTML]{009901} +0.57}  & {\color[HTML]{009901} +0.45}     & {\color[HTML]{CB0000} -0.17}  & {\color[HTML]{009901} +1.17}     & {\color[HTML]{009901} +0.49}    & {\color[HTML]{009901} +0.49}  & {\color[HTML]{CB0000} -0.08}     & {\color[HTML]{009901} +0.97}  & {\color[HTML]{009901} +0.85}     \\ \cmidrule{1-1}\cmidrule{3-14}
\textbf{CLIP-Adapter} \cite{gao2023clip}&      & 62.83                             & 73.25                       & 62.12                        & 23.85                        & 80.43                            & 65.18                         & 80.93                            & 67.97                          & 88.28                         & 92.09                            & 78.23                         & 70.46                          \\
$\bm\Delta \textbf{+KCL}$(Ours) &       & {\color[HTML]{009901} +0.88}      & {\color[HTML]{009901} +2.14} & {\color[HTML]{009901} +1.18} & {\color[HTML]{009901} +3.60}  & {\color[HTML]{009901} +9.83}     & {\color[HTML]{009901} +1.38}  & {\color[HTML]{009901} +1.17}     & {\color[HTML]{009901} +0.43}    & {\color[HTML]{009901} +0.22}  & {\color[HTML]{009901} +0.24}     & {\color[HTML]{009901} +0.40}  & {\color[HTML]{009901} +1.95}     \\ \cmidrule{1-1}\cmidrule{3-14}
\textbf{Tip-Adapter} \cite{zhang2022tip} &      & 62.42                             & 70.58                        & 60.99                        & 27.69                         & 87.66                            & 64.68                         & 69.85                            & 66.74                           & 86.45                         & 91.08                            & 77.72                          & 69.62                           \\
$\bm\Delta \textbf{+KCL}$(Ours) &       & {\color[HTML]{009901} +0.49}      & {\color[HTML]{009901} +2.75} & {\color[HTML]{009901} +1.01} & {\color[HTML]{009901} +0.48}  & {\color[HTML]{009901} +1.26}     & {\color[HTML]{CB0000} -0.12}   & {\color[HTML]{009901} +2.78}      & {\color[HTML]{009901} +0.07}    & {\color[HTML]{009901} +0.47}  & {\color[HTML]{CB0000} -0.37}     & {\color[HTML]{009901} +0.40}  & {\color[HTML]{009901} +0.83}     \\ \cmidrule{1-1}\cmidrule{3-14}

\textbf{APE} \cite{zhu2023not} &      & 62.76                            & 71.64                       & 64.78                        & 28.86                         & 90.35                           & 66.74                          &  72.60                           & 68.01                           & 86.63                         & 91.08                           & 78.10                         &  71.05                            \\
$\bm\Delta \textbf{+KCL}$(Ours) &       & {\color[HTML]{009901} +0.85}      & {\color[HTML]{009901} +4.81} & {\color[HTML]{009901} +0.59} & {\color[HTML]{009901} +0.09}  & {\color[HTML]{009901} +0.72}     & {\color[HTML]{009901} +0.34}  & {\color[HTML]{009901} +2.51}     & {\color[HTML]{CB0000} -0.29}     & {\color[HTML]{009901} +0.75}   & {\color[HTML]{009901} +0.48}     & {\color[HTML]{009901} +0.16}  & {\color[HTML]{009901} +1.00}     \\ \cmidrule{1-1}\cmidrule{3-14}

\textbf{MaPLe} \cite{khattak2023maple} &      & 71.42                             & 79.25                        & 62.23                        & 32.31                         & 90.38                            & 71.27                         & 76.05                            & 72.61                           & 93.40                         & 94.60                            & 87.17                         & 75.51                           \\
$\bm\Delta \textbf{+KCL}$(Ours) &       & {\color[HTML]{009901} +1.11}      & {\color[HTML]{009901} +4.89} & {\color[HTML]{009901} +5.80} & {\color[HTML]{009901} +8.97}  & {\color[HTML]{009901} +4.34}    & {\color[HTML]{009901} +4.74}   & {\color[HTML]{009901} +9.76}    & {\color[HTML]{009901} +2.43}    & {\color[HTML]{009901} +0.11}  & {\color[HTML]{009901} +0.45}     & {\color[HTML]{009901} +0.13}  & {\color[HTML]{009901} +3.88}     \\ \midrule

\textbf{CoOp} \cite{zhou2022learning} & \multirow{10}{*}{\textbf{16}} & 61.12                             & 78.61                        & 68.68                        & 36.03                         & 95.66                            & 76.26                         & 85.58                             & 70.02                           & 88.25                         & 93.67                            & 76.91                         & 75.52                           \\
$\bm\Delta \textbf{+KCL}$(Ours) &       & {\color[HTML]{009901} +0.77}      & {\color[HTML]{009901} +1.25} & {\color[HTML]{009901} +0.17} & {\color[HTML]{009901} +0.18}                             & {\color[HTML]{009901} +0.04}      & {\color[HTML]{CB0000} -0.24}  & {\color[HTML]{009901} +0.21}     & {\color[HTML]{009901} +0.48}    & {\color[HTML]{009901} +0.52}  & {\color[HTML]{CB0000} -0.24}     & {\color[HTML]{009901} +0.86}   & {\color[HTML]{009901} +0.36}     \\ \cmidrule{1-1}\cmidrule{3-14}
\textbf{CLIP-Adapter} \cite{gao2023clip}&      & 63.24                            & 76.53                        & 66.55                        & 29.01                          & 92.94                            & 71.51                         & 84.67                            & 69.79                           & 88.53                         & 92.66                            & 78.59                         & 74.00                          \\
$\bm\Delta \textbf{+KCL}$(Ours) &       & {\color[HTML]{009901} +0.93}      & {\color[HTML]{009901} +4.07} & {\color[HTML]{009901} +0.03} & {\color[HTML]{009901} +4.02}  & {\color[HTML]{009901} +0.73}     & {\color[HTML]{009901} +0.89}  & {\color[HTML]{009901} +0.53}     & {\color[HTML]{009901} +0.46}    & {\color[HTML]{009901} +0.19}  & 0.00     & {\color[HTML]{009901} +0.49}  & {\color[HTML]{009901} +1.12}     \\ \cmidrule{1-1}\cmidrule{3-14}
\textbf{Tip-Adapter} \cite{zhang2022tip} &      & 63.00                             & 72.16                        & 63.42                        & 30.75                         & 89.81                             & 66.73                         & 74.83                            & 68.13                            & 86.48                         & 91.08                            & 77.81                         & 71.29                          \\
$\bm\Delta \textbf{+KCL}$(Ours) &       & {\color[HTML]{009901} +0.04}      & {\color[HTML]{009901} +1.72} & {\color[HTML]{009901} +0.82}  & {\color[HTML]{009901} +0.03}  & {\color[HTML]{009901} +0.24}     & {\color[HTML]{009901} +1.04}  & {\color[HTML]{CB0000} --0.05}     & {\color[HTML]{009901} +1.04}    & {\color[HTML]{009901} +0.96}  & {\color[HTML]{CB0000} -0.25}     & {\color[HTML]{009901} +0.14}  & {\color[HTML]{009901} +0.52}     \\ \cmidrule{1-1}\cmidrule{3-14}

\textbf{APE} \cite{zhu2023not} &      & 63.45                            & 74.57                       & 66.19                        & 31.80                         & 92.00                           & 70.48                          &  77.54                          & 69.30                           & 87.23                         & 91.68                           & 78.44                         &  72.94                           \\
$\bm\Delta \textbf{+KCL}$(Ours) &       & {\color[HTML]{009901} +0.40}      & {\color[HTML]{009901} +3.20} & {\color[HTML]{009901} +1.30} & {\color[HTML]{009901} +0.18}  & {\color[HTML]{009901} +0.08}     & {\color[HTML]{009901} +0.12}  & {\color[HTML]{009901} +1.68}     & {\color[HTML]{009901} +0.13}     & {\color[HTML]{CB0000} -0.42}   & {\color[HTML]{009901} +0.04}     & {\color[HTML]{CB0000} -0.22}  & {\color[HTML]{009901} +0.59}     \\ \cmidrule{1-1}\cmidrule{3-14}

\textbf{MaPLe} \cite{khattak2023maple} &      & 71.83                             & 81.10                        & 67.91                        & 35.55                          & 93.59                            & 73.15                         & 86.02                             & 74.56                              & 94.22                         & 95.54                            & 87.27                         & 78.24                           \\
$\bm\Delta \textbf{+KCL}$(Ours) &       & {\color[HTML]{009901} +1.58}      & {\color[HTML]{009901} +2.99} & {\color[HTML]{009901} +5.20} & {\color[HTML]{009901} +9.39}  & {\color[HTML]{009901} +1.86}     & {\color[HTML]{009901} +6.93}  & {\color[HTML]{009901} +4.34}    & {\color[HTML]{009901} +1.83}    & {\color[HTML]{CB0000} -0.57}  & {\color[HTML]{009901} +0.16}     & {\color[HTML]{009901} +0.27}  & {\color[HTML]{009901} +3.08}     \\
\bottomrule[1pt]
\end{tabular}
}
\end{table}
\newpage

\section{Ablation Study on Visual Backbones}%
In this section, we conduct ablation study on four different visual backbone structures, i.e. ResNet-50 \cite{he2016deep}, ResNet-101 \cite{he2016deep}, ViT-B/32 \cite{dosovitskiy2020image} and ViT-B/16 \cite{dosovitskiy2020image}. As is shown in Table \ref{backbone}, we present the full performance with different backbone structures under 1, 2, 4, 8 and 16-shot settings. We can find that KCL still achieves remarkable improvements based on various backbones.

\begin{table}[h]
\centering
\caption{Ablation study of visual backbones on zero-shot CLIP \cite{radford2021learning} and CoOp \cite{zhou2022learning}.}
\label{backbone}
\resizebox{0.98\textwidth}{!}{
\begin{tabular}{lc|cccc|cccc}
\toprule[1pt]

\multicolumn{10}{c}{\textbf{Zero-shot}}\\ \midrule
\multicolumn{2}{l|}{\textbf{Methods}} & \multicolumn{4}{c|}{\textbf{CLIP}}                             & \multicolumn{4}{c}{\textbf{CLIP+KCL}}    \\ \midrule 
\textbf{Datasets} & \textbf{Shots}  & \textbf{RN50}  & \textbf{RN101} & \textbf{ViT/16} & \textbf{ViT/32} & \textbf{RN50}  & \textbf{RN101} & \textbf{ViT/16} & \textbf{ViT/32} \\ \midrule

\textbf{UCF}                                         &                                           0& 59.69 & 55.46 & 67.06  & \multicolumn{1}{c|}{63.57}  & 67.06 & 63.23 & 72.98  & 67.88  \\ \midrule
\textbf{Eurosat}                                     &                  0& 38.02 & 30.51 & 54.16  & \multicolumn{1}{c|}{48.30}  & 52.69 & 33.36 & 56.64  & 50.56  \\ \midrule
\textbf{Cars}                                        &                      0& 57.34 & 61.41 & 65.99  & \multicolumn{1}{c|}{61.09}  & 59.68 & 65.08 & 66.48  & 61.73  \\ \midrule
\textbf{DTD}                                      &     0  & 48.52 & 49.88 & 53.31  & \multicolumn{1}{c|}{50.06}  & 56.21    & 53.31 & 56.80  & 54.79 \\ \midrule
\multicolumn{10}{c}{\textbf{Few-shot}}\\ \midrule
\multicolumn{2}{l|}{\textbf{Methods}} & \multicolumn{4}{c|}{\textbf{CoOp}}                             & \multicolumn{4}{c}{\textbf{CoOp+KCL}}    \\ \midrule 
\textbf{Datasets} & \textbf{Shots}  & \textbf{RN50}  & \textbf{RN101} & \textbf{ViT/16} & \textbf{ViT/32} & \textbf{RN50}  & \textbf{RN101} & \textbf{ViT/16} & \textbf{ViT/32} \\ \midrule

\multirow{5}{*}{\textbf{UCF}}     & 1                     & 64.92 & 67.22 & 72.83  & 70.02                       & 69.65 & 73.22 & 80.28  & 77.21  \\ 
                         & 2                     & 68.44 & 72.91 & 76.92  & 74.12                       & 73.41 & 75.63 & 81.55  & 76.29  \\ 
                         & 4                     & 70.71 & 75.31 & 79.80  & 77.43                       & 74.99 & 78.19 & 82.13  & 80.44  \\ 
                         & 8                     & 74.76 & 79.65 & 83.16  & 80.49                       & 78.61 & 81.26 & 84.96  & 82.66  \\ 
                         & 16                    & 78.61 & 81.34 & 84.99  & 82.87                       & 79.86 & 82.18 & 85.75  & 83.40  \\ \midrule
\multirow{5}{*}{\textbf{Eurosat}} & 1                     & 62.56 & 57.17 & 58.80  & 61.44                       & 69.31 & 63.70 & 65.44  & 69.67  \\  
                         & 2                     & 65.49 & 64.30 & 69.54  & 67.79                       & 73.01 & 72.36 & 79.91  & 76.00  \\ 
                         & 4                     & 72.70  & 73.62 & 82.17  & 80.42                       & 76.99 & 76.70 & 84.64  & 81.86  \\ 
                         & 8                     & 78.60  & 78.20 & 86.05  & 81.70                       & 79.77 & 78.44 & 87.09  & 82.10  \\  
                         & 16                    & 85.58 & 80.52 & 87.48  & 84.17                       & 85.79 & 82.19 & 87.78  & 84.96  \\ \midrule
\multirow{5}{*}{\textbf{Cars}}    & 1                     & 53.89 & 63.57 & 66.83  & 59.05                       & 57.29 & 64.84 & 67.55  & 61.25  \\ 
                         & 2                     & 60.17 & 68.00 & 71.17  & 63.09                       & 61.20  & 68.74 & 71.57  & 63.54  \\ 
                         & 4                     & 64.71 & 71.05 & 74.74  & 66.55                       & 64.90  & 71.30 & 74.88  & 67.03  \\ 
                         & 8                     & 69.83 & 76.76 & 80.70  & 73.44                       & 69.66 & 76.87 & 80.74  & 73.47  \\ 
                         & 16                    & 76.26 & 81.68 & 83.36  & 78.61                       & 76.02 & 81.71 & 83.67  & 78.77  \\ \midrule
\multirow{5}{*}{\textbf{DTD}} & 1                        & 53.25 & 55.50 & 57.74  & 55.44                       & 57.98    & 58.22 & 60.93  & 57.86  \\  
                    & 2                        & 55.85 & 58.98 & 60.64  & 58.27                       & 56.38    & 59.34 & 62.94  & 62.53  \\  
                    & 4                        & 61.64 & 63.06 & 64.72  & 63.30                       & 61.82    & 64.01 & 65.84  & 65.13  \\  
                    & 8                        & 64.78 & 67.55 & 70.39  & 66.55                       & 65.43    & 67.73 & 69.27  & 66.55  \\  
                    & 16                       & 68.68 & 70.33 & 73.17  & 69.74                       & 68.85    & 70.33 & 73.23  & 69.33  \\

                         \bottomrule[1pt]
\end{tabular}
}
\end{table}

\newpage
\section{Analysis of Knowledge Completion Capability}
\par
\begin{figure}[h]
  \centering
  \begin{subfigure}{0.32\linewidth}
    \includegraphics[width=0.95\linewidth]{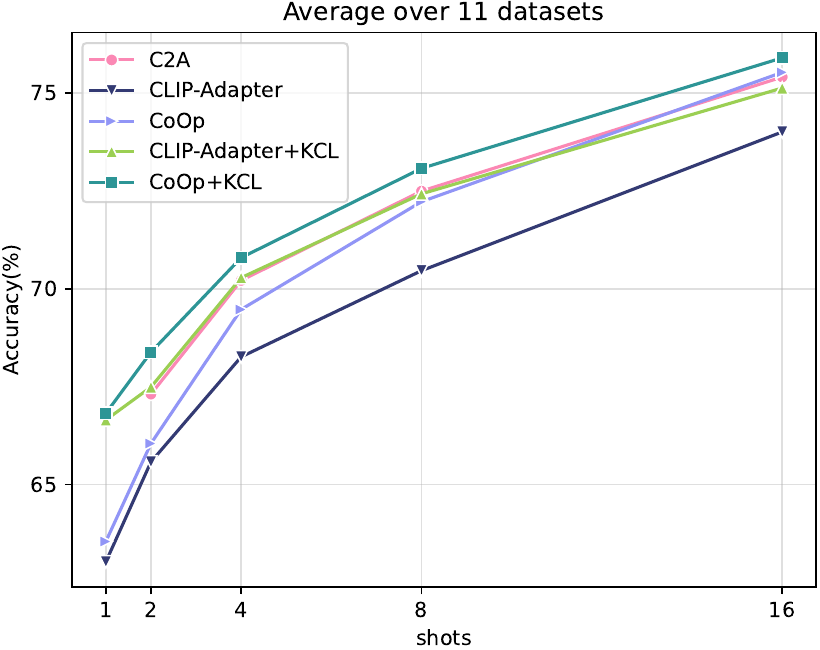}
    \label{fig:2-a}
  \end{subfigure}
  \begin{subfigure}{0.32\linewidth}
    \includegraphics[width=0.95\linewidth]{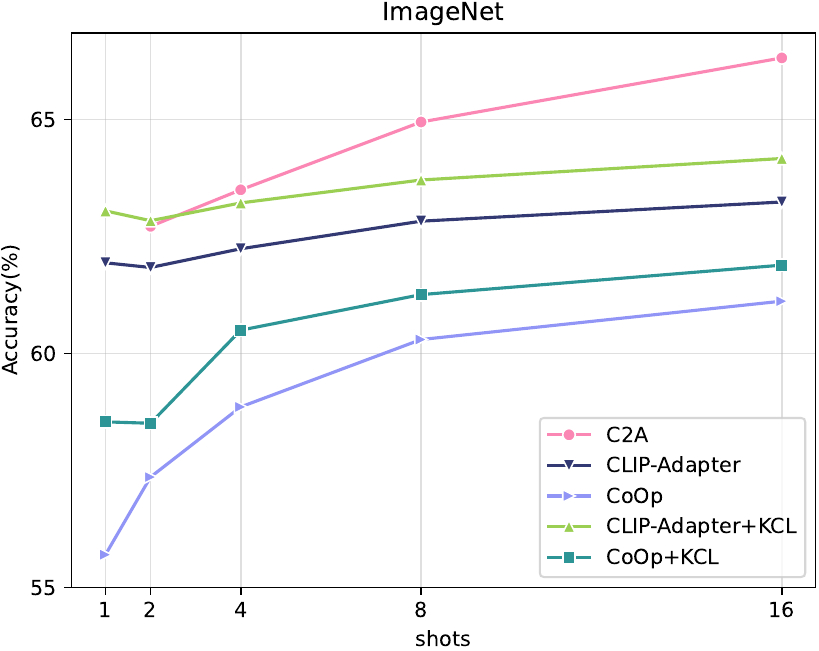}
    \label{fig:2-b}
  \end{subfigure}
   \begin{subfigure}{0.32\linewidth}
    \includegraphics[width=0.95\linewidth]{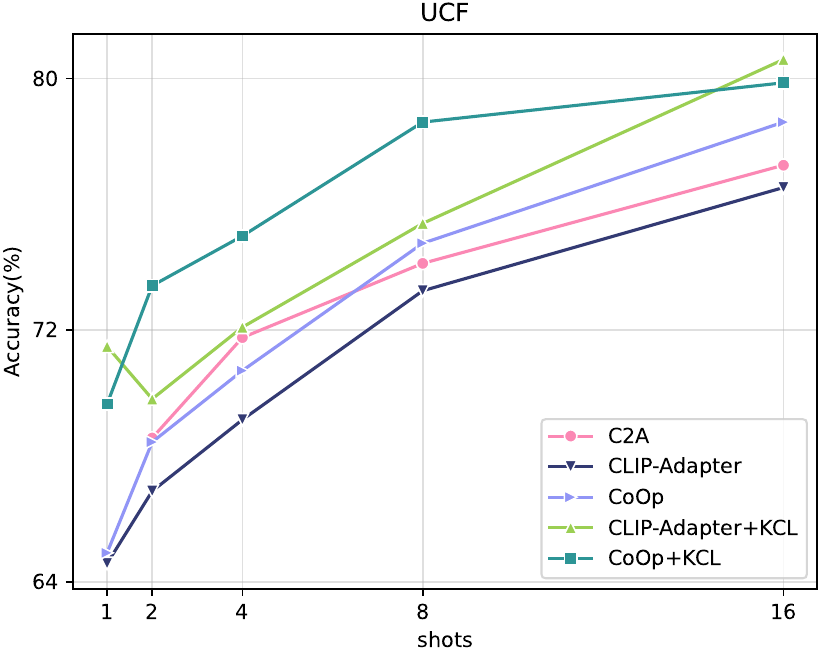}
    \label{fig:2-c}
  \end{subfigure}
  \begin{subfigure}{0.32\linewidth}
    \includegraphics[width=0.95\linewidth]{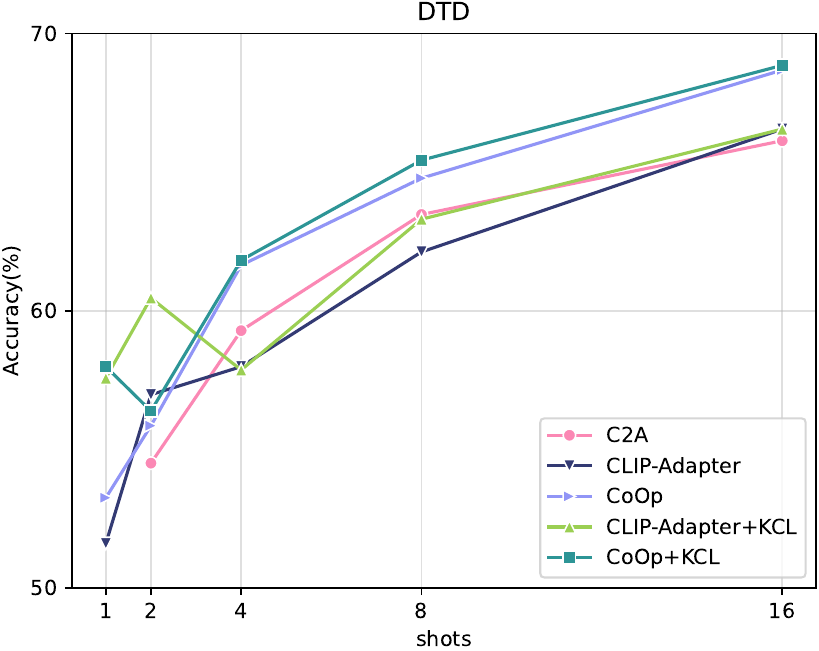}
    \label{fig:2-e}
  \end{subfigure}
    \begin{subfigure}{0.32\linewidth}
    \includegraphics[width=0.95\linewidth]{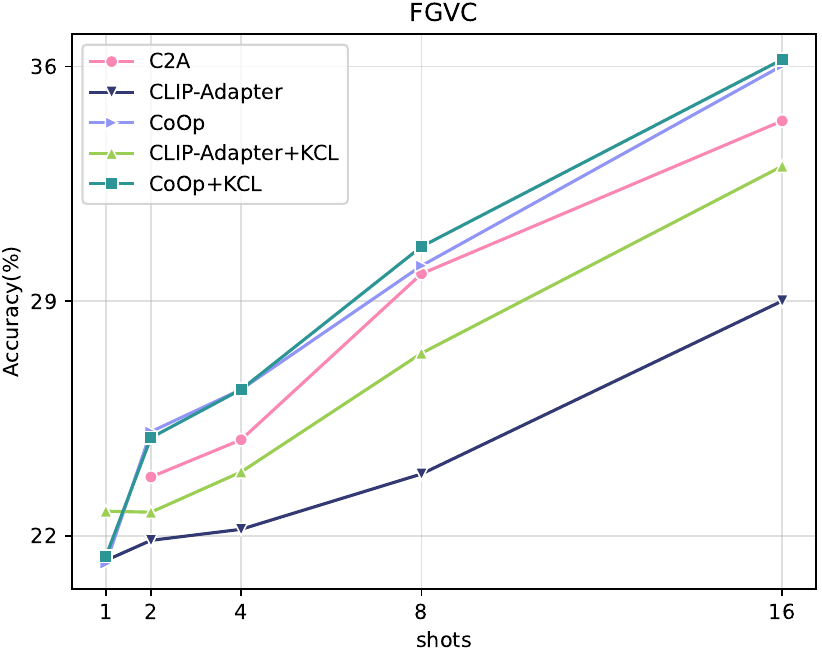}
    \label{fig:2-f}
  \end{subfigure}
    \begin{subfigure}{0.32\linewidth}
    \includegraphics[width=0.95\linewidth]{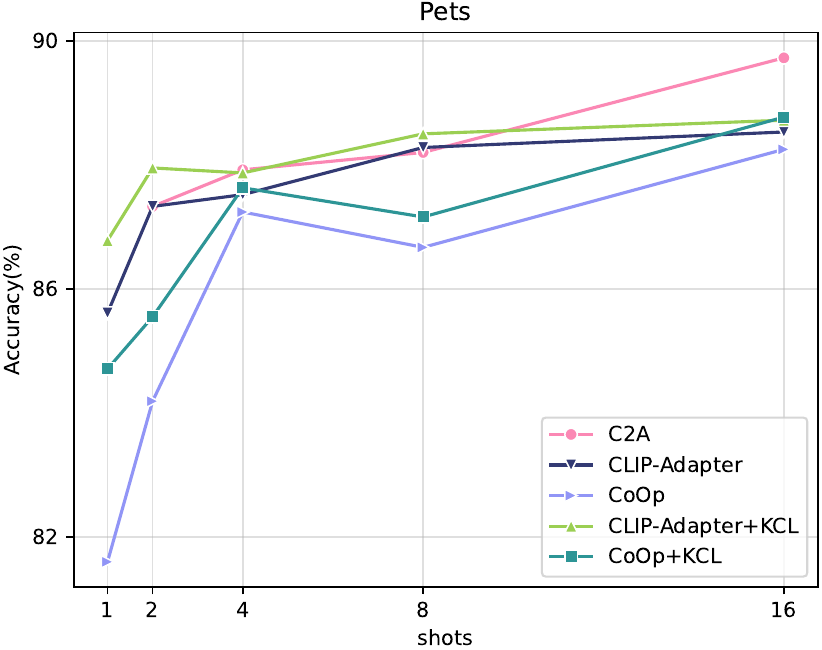}
    \label{fig:2-d}
  \end{subfigure}
  \begin{subfigure}{0.32\linewidth}
    \includegraphics[width=0.95\linewidth]{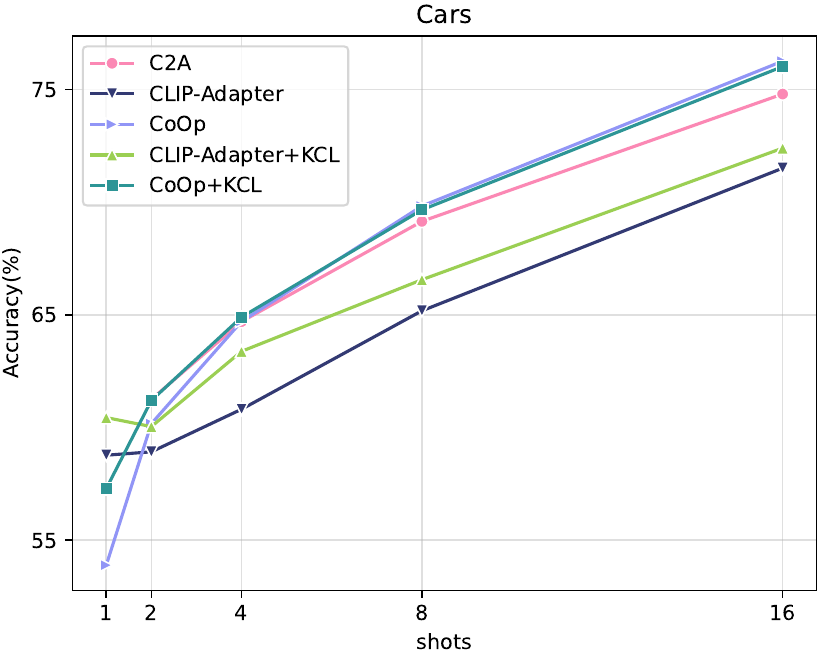}
    \label{fig:2-a}
  \end{subfigure}
  \begin{subfigure}{0.32\linewidth}
    \includegraphics[width=0.95\linewidth]{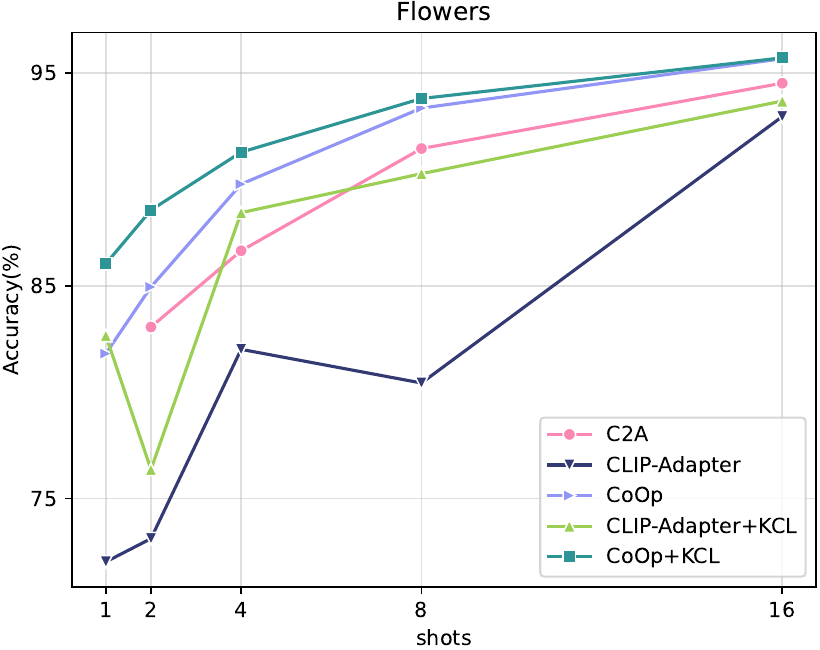}
    \label{fig:2-b}
  \end{subfigure}
   \begin{subfigure}{0.32\linewidth}
    \includegraphics[width=0.95\linewidth]{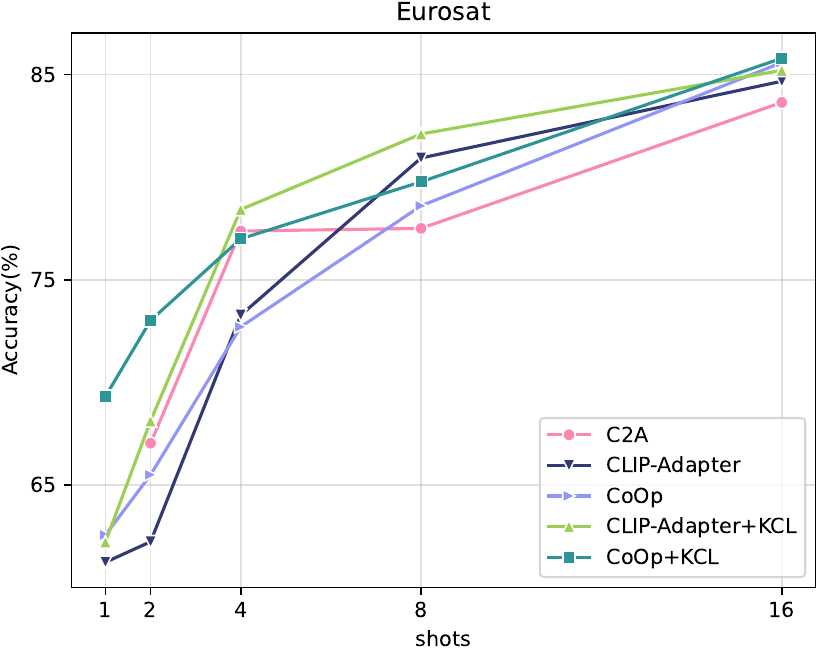}
    \label{fig:2-c}
  \end{subfigure}
  \begin{subfigure}{0.32\linewidth}
    \includegraphics[width=0.95\linewidth]{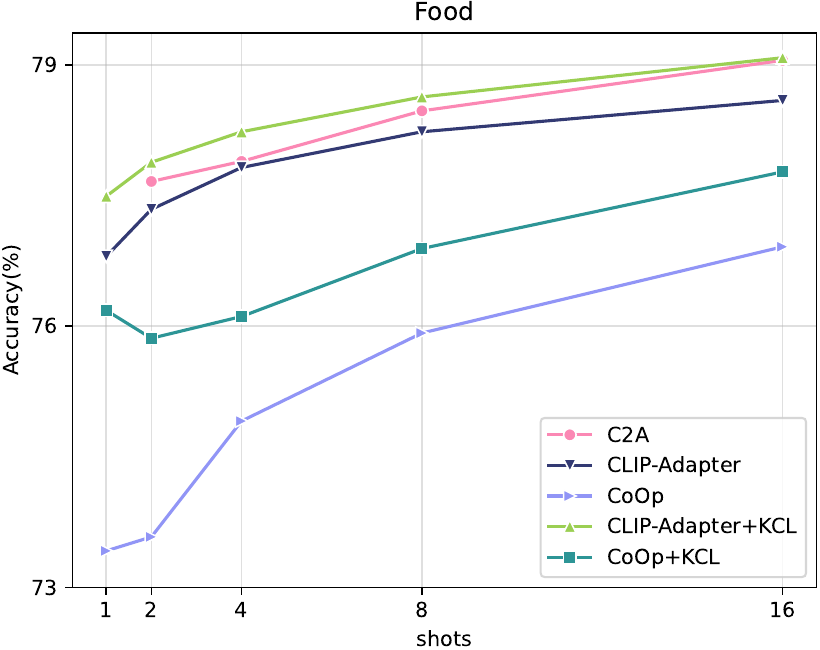}
    \label{fig:2-e}
  \end{subfigure}
    \begin{subfigure}{0.32\linewidth}
    \includegraphics[width=0.95\linewidth]{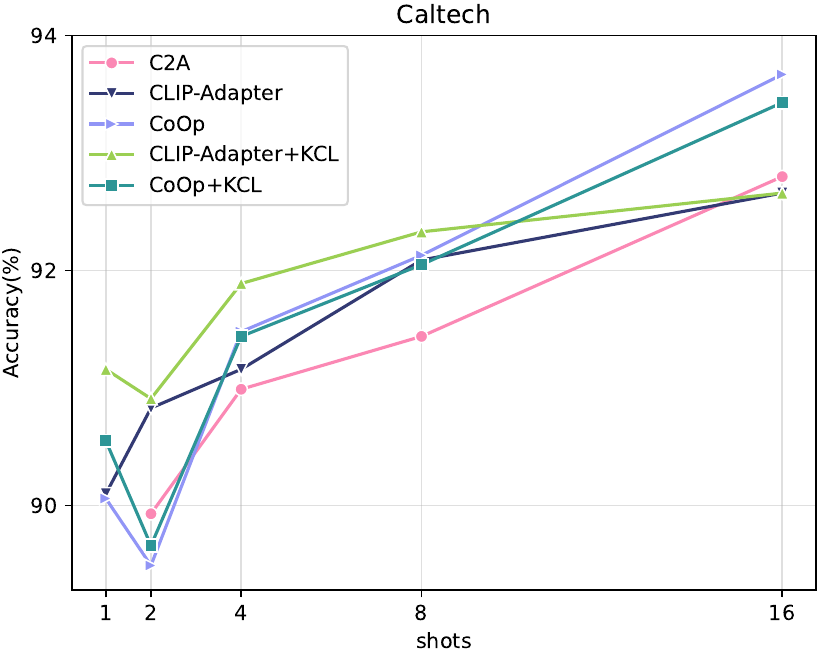}
    \label{fig:2-f}
  \end{subfigure}
    \begin{subfigure}{0.32\linewidth}
    \includegraphics[width=0.95\linewidth]{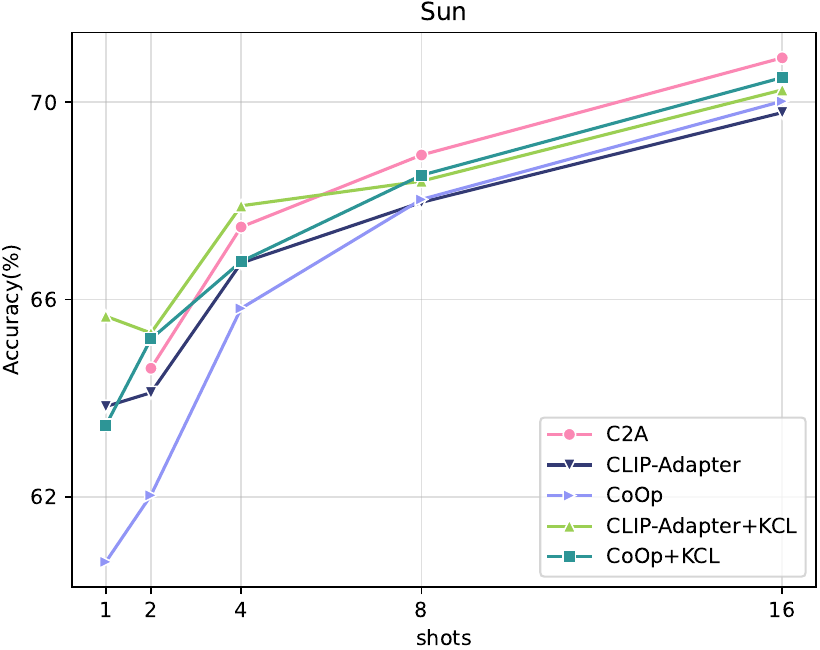}
    \label{fig:2-d}
  \end{subfigure}
  \hfill
  \caption{Knowledge completion capability comparison with C2A \cite{roy2023Cap2Aug}.}
  \label{sp.fig:c2a}
\end{figure}

 In Fig. 2a of main paper, we compare the knowledge completion performance of KCL with generation-based C2A \cite{roy2023Cap2Aug}. C2A, building up on the combination of CLIP-Adapter \cite{gao2023clip} and Tip-Adapter \cite{zhang2022tip}, takes advantages of stable diffusion \cite{rombach2022high} and BLIP \cite{li2022blip} to generate images as visual knowledge completion. Here we present the overall results on 11 datasets in Fig. \ref{sp.fig:c2a}. We can find that KCL module embedded in CLIP-Adapter can achieve competitive performance with C2A, while KCL module embedded in CoOp can surpass C2A in most cases. Note that KCL has no need for generative models with high computational costs.

\newpage
\section{Hyper-parameter Sensitivity Analysis}
\begin{figure}[htbp]
  \centering
  \begin{subfigure}{0.32\linewidth}
    \includegraphics[width=0.95\linewidth]{figures/fix/fix_avg.pdf}
    \label{fig:1-a}
  \end{subfigure}
  \begin{subfigure}{0.32\linewidth}
    \includegraphics[width=0.95\linewidth]{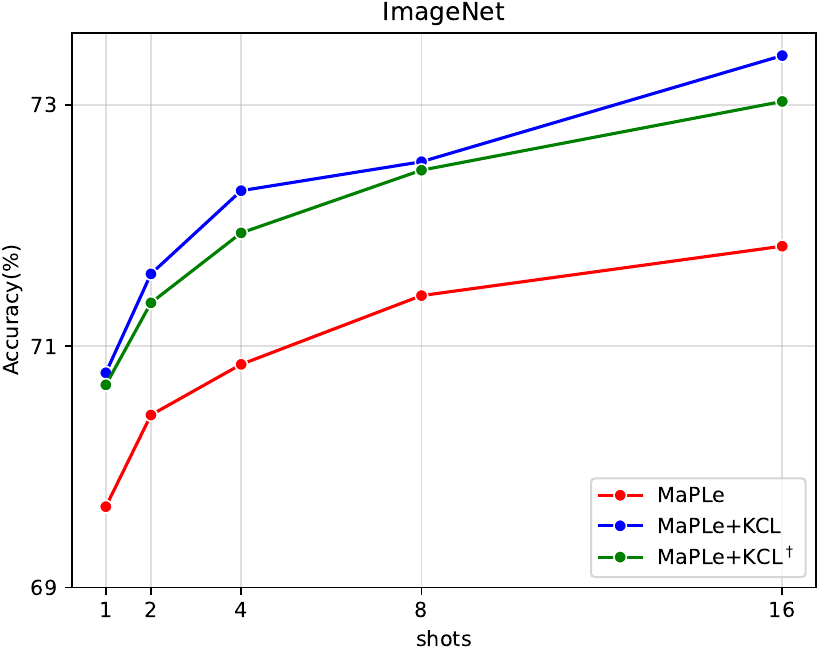}
    \label{fig:1-b}
  \end{subfigure}
   \begin{subfigure}{0.32\linewidth}
    \includegraphics[width=0.95\linewidth]{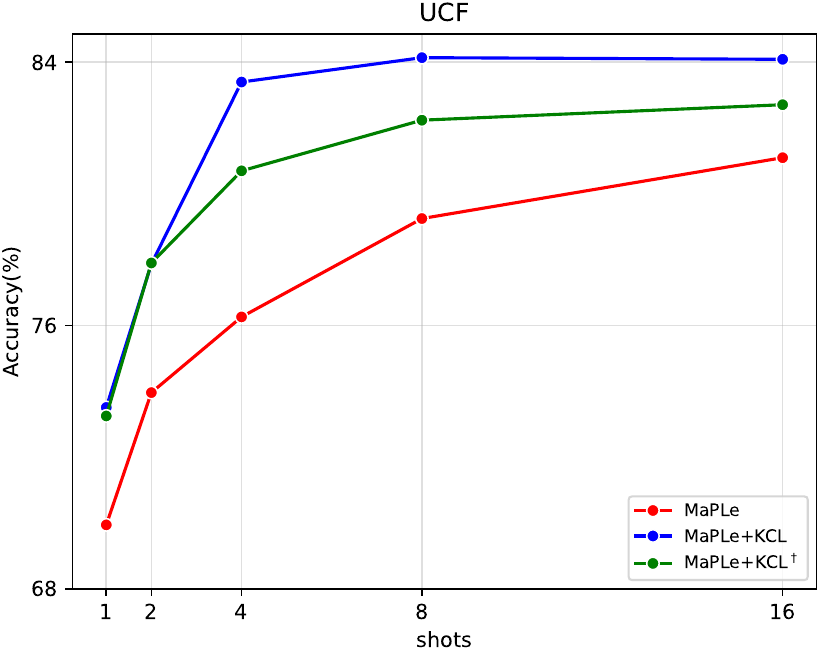}
    \label{fig:1-c}
  \end{subfigure}
  \begin{subfigure}{0.32\linewidth}
    \includegraphics[width=0.95\linewidth]{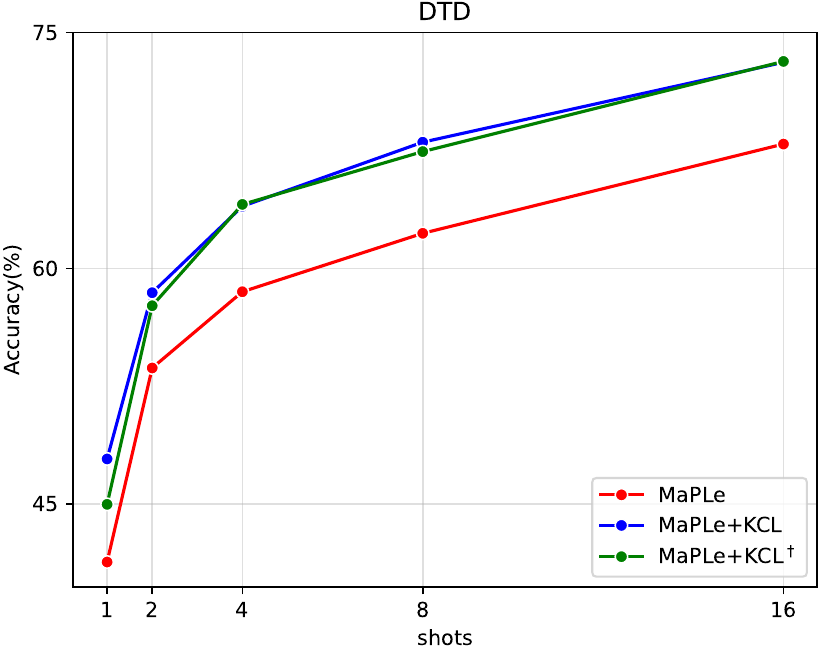}
    \label{fig:1-e}
  \end{subfigure}
    \begin{subfigure}{0.32\linewidth}
    \includegraphics[width=0.95\linewidth]{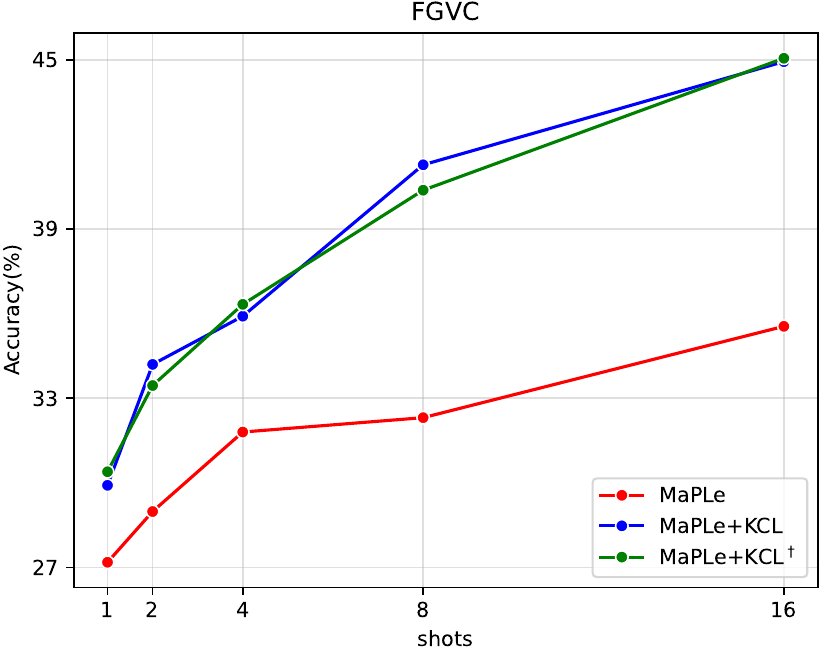}
    \label{fig:1-f}
  \end{subfigure}
    \begin{subfigure}{0.32\linewidth}
    \includegraphics[width=0.95\linewidth]{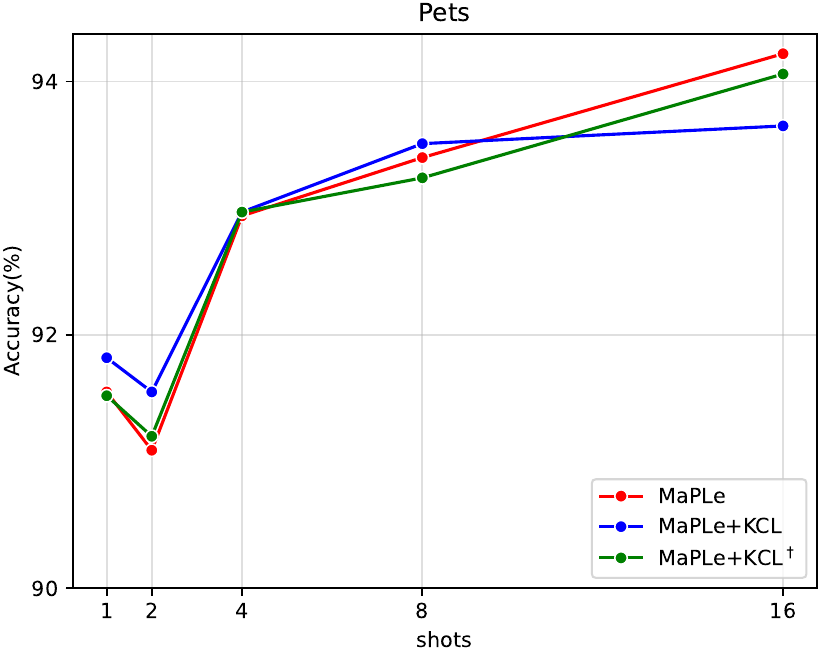}
    \label{fig:1-d}
  \end{subfigure}
  \begin{subfigure}{0.32\linewidth}
    \includegraphics[width=0.95\linewidth]{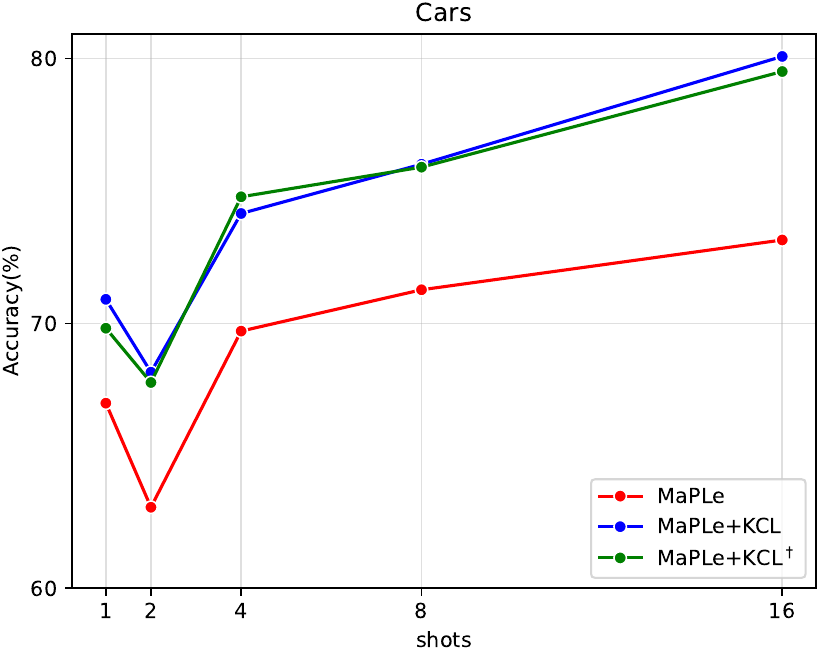}
    \label{fig:1-a}
  \end{subfigure}
  \begin{subfigure}{0.32\linewidth}
    \includegraphics[width=0.95\linewidth]{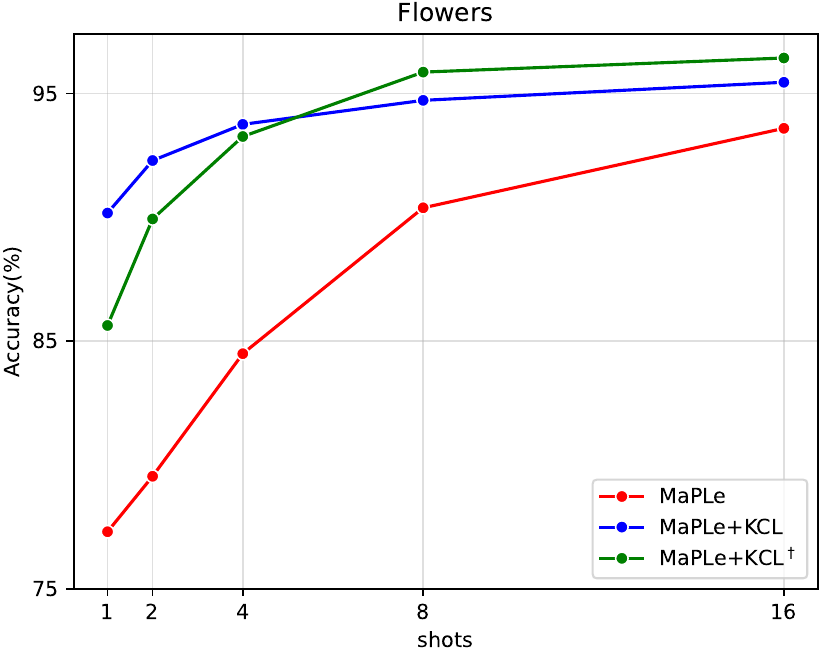}
    \label{fig:1-b}
  \end{subfigure}
   \begin{subfigure}{0.32\linewidth}
    \includegraphics[width=0.95\linewidth]{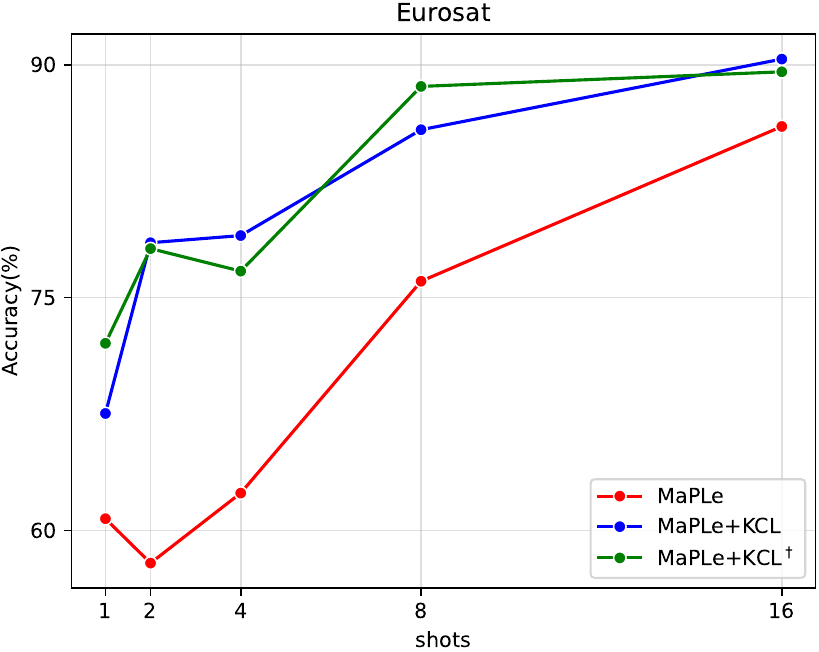}
    \label{fig:1-c}
  \end{subfigure}
  \begin{subfigure}{0.32\linewidth}
    \includegraphics[width=0.95\linewidth]{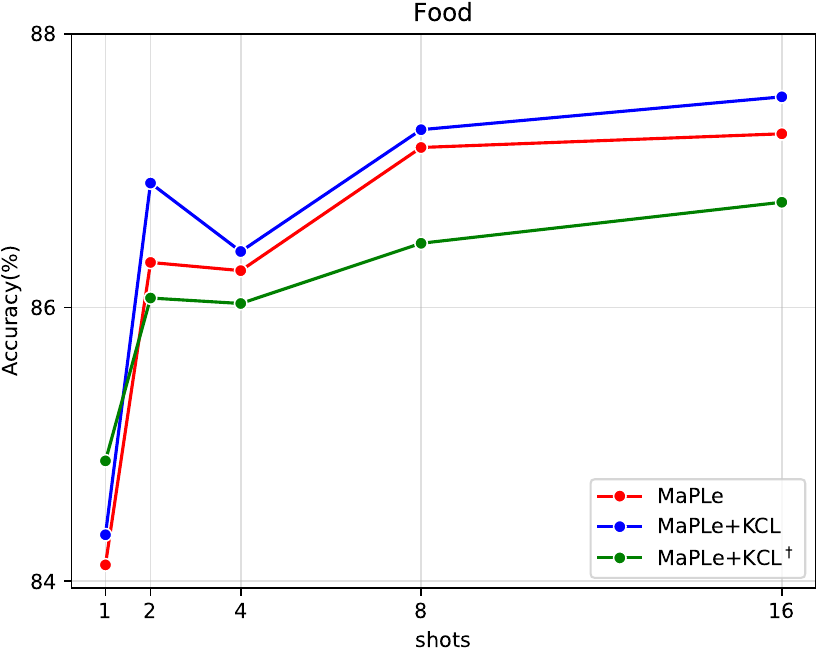}
    \label{fig:1-e}
  \end{subfigure}
    \begin{subfigure}{0.32\linewidth}
    \includegraphics[width=0.95\linewidth]{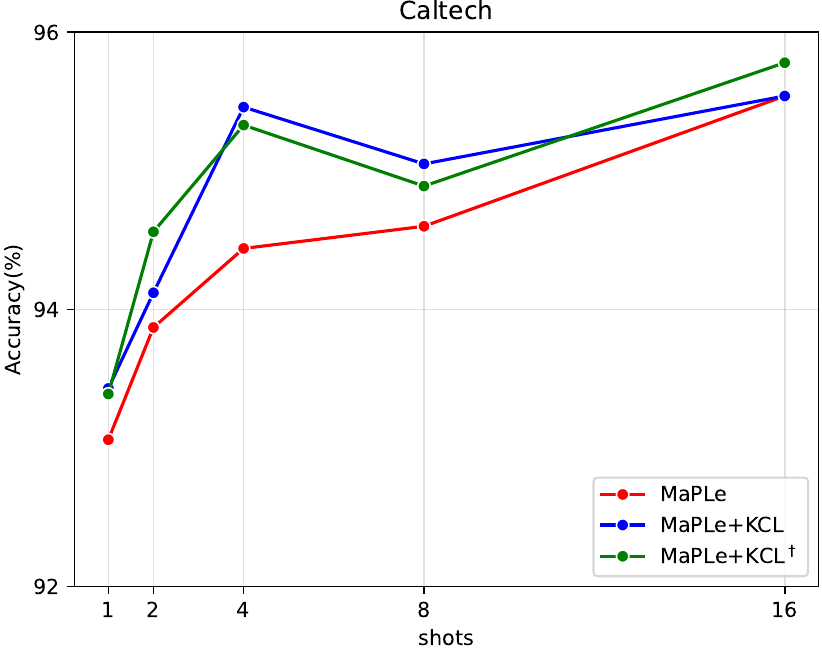}
    \label{fig:1-f}
  \end{subfigure}
    \begin{subfigure}{0.32\linewidth}
    \includegraphics[width=0.95\linewidth]{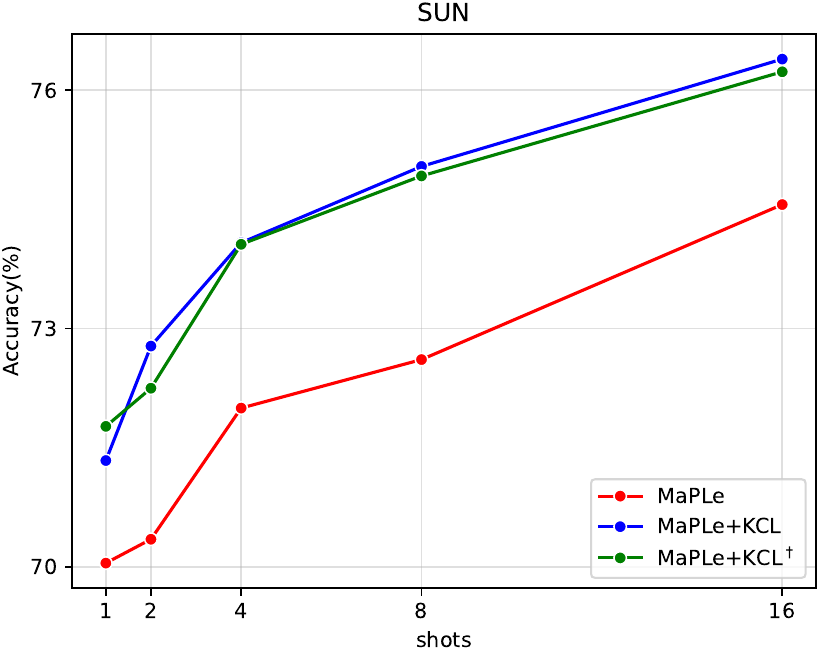}
    \label{fig:1-d}
  \end{subfigure}
  \hfill
  \caption{Overall experimental results of hyper-parameter sensitivity analysis.}
  \label{sp:fig:2}
\end{figure}

In Fig. 2b of main paper, we show the averaged results of hyper-parameter sensitivity analysis. In this section, we show the results of on 11 datasets based on MaPLe in \cref{sp:fig:2}. $\dagger$ means all hyper-parameters are fixed to $\alpha$=$\beta$=$\mu$=$\lambda$=1.0 without tuning hyper-parameters. We can find that hyper-parameters have tiny impacts towards KCL, which demonstrates the robustness of KCL.

\section{Textual Prompts Provided by CuPL}
The textual prompts used in this work is provided by CuPL \cite{pratt2023does}, which is generated by querying GPT-3 \cite{dale2021gpt}.
\begin{table}[htbp]
    \centering
    \caption{\textbf{CuPL \cite{pratt2023does} hand-written prompts.}}  
    \resizebox{0.9\textwidth}{!}{
    \begin{tabular}{ll}
    \toprule[1pt]
        \textbf{Dataset} & \textbf{GPT-3 prompts} \\
        \midrule
        \multirow{5}{*}{}{\textbf{ImageNet}} & `Describe what a \{\} looks like'\\
	&	`How can you identify \{\}?'\\
       & `What does \{\} look like?'\\
	&	`Describe an image from the internet of a \{\}'\\
	& `A caption of an image of \{\}: '\\
  \midrule
        \multirow{3}{*}{}{\textbf{UCF}} & `What does a person doing \{\} look like'\\
        & `Describe the process of \{\}'\\
		& `How does a person \{\}' \\
  \midrule
        \multirow{6}{*}{}{\textbf{DTD}} & `What does a \{\} material look like?'\\
	&	`What does a \{\} surface look like?'\\
	&	`What does a \{\} texture look like?'\\
	&	`What does a \{\} object look like?'\\
	&	`What does a \{\} thing look like?'\\
	&	`What does a \{\} pattern look like?'\\
 \midrule

        \multirow{1}{*}{}{\textbf{FGVC}} & `Describe a \{\} aircraft'\\
  \midrule

        \multirow{4}{*}{}{\textbf{Flowers}} & `What does a \{\} flower look like'\\
		& `Describe the appearance of a \{\}'\\
		& `A caption of an image of \{\}'\\
		& `Visually describe a \{\}, a type of flower'\\
  \midrule

        \multirow{9}{*}{}{\textbf{Cars}} & `How can you identify a \{\}'\\
		& `Description of a \{\}, a type of car'\\
		& `A caption of a photo of a \{\}:'\\
		& `What are the primary characteristics of a \{\}?'\\
		& `Description of the exterior of a \{\}'\\
		& `What are the identifying characteristics of a \{\}, a type of car?'\\
		& `Describe an image from the internet of a \{\}'\\
		& `What does a \{\} look like?'\\
		& `Describe what a \{\}, a type of car, looks like'\\
  \bottomrule

        \multirow{3}{*}{}{\textbf{Eurosat}} & `Describe an aerial satellite view of \{\}'\\
	&	`How does a satellite photo of a \{\} look like'\\
	&	`Visually describe a centered satellite view of a \{\}'\\
 \midrule

        \multirow{3}{*}{}{\textbf{SUN}} & `Describe what a \{\} looks like'\\
		& `How can you identify a \{\}?'\\
		& `Describe a photo of a \{\}'\\
  \midrule

        \multirow{2}{*}{}{\textbf{Pets}} & `Describe what a \{\} pet looks like'\\
 & `Visually describe a \{\}, a type of pet'\\
 \midrule
        
        \multirow{3}{*}{}{\textbf{Caltech}} & `Describe what a \{\} looks like'\\
		& `What does a \{\} look like'\\
	&	`Describe a photo of a \{\}'\\
  \midrule

        \multirow{3}{*}{}{\textbf{Food}} & `Describe what a \{\} looks like'\\
		& `Visually describe a \{\}'\\
	& `How can you tell that the food in this photo is a \{\}?'\\
 \bottomrule[1pt]
        
\end{tabular}
}
    \label{tab:cupl-prompts-1}
\end{table}

\newpage

\bibliographystyle{splncs04}
\bibliography{egbib}

\end{document}